%% file: main.tex
\documentclass[runningheads]{llncs}

\usepackage{eccv}

\usepackage{eccvabbrv}

\usepackage{graphicx}
\usepackage{booktabs}
\usepackage{bbm}

\usepackage[accsupp]{axessibility}  % Improves PDF readability for those with disabilities.

\usepackage{hyperref}

\usepackage{orcidlink}
\usepackage[dvipsnames]{xcolor}
\usepackage{booktabs}
\usepackage{multirow} 
\RequirePackage{enumitem}
\setlist[itemize]{noitemsep,leftmargin=*,topsep=0em}
\setlist[enumerate]{noitemsep,leftmargin=*,topsep=0em}
\usepackage{array}
\newcolumntype{C}[1]{p{#1}}
\usepackage{icomma}
\begin{document}

\title{Weighting Pseudo-Labels via High-Activation Feature Index Similarity and Object Detection for Semi-Supervised Segmentation} 

\titlerunning{Weighting Pseudo-Labels}


\author{Prantik Howlader\inst{1} \and
Hieu Le\inst{2} \and
Dimitris Samaras\inst{1}}

\authorrunning{P. Howlader et al.}

% \institute{Stonybrook University \and
% \email{\{phowlader; samaras\}@cs.stonybrook.edu}\\

\institute{Stony Brook University, New York, USA \\
%\email{\{phowlader; samaras\}@cs.stonybrook.edu} 
\and
EPFL, Lausanne, Switzerland\\
%\email{minh.le@epfl.ch}
}
\maketitle

%\textcolor{red}{HIEU:Correcting pseudo-labels via  bounding box classifier and high-activate indexes similarity.}
\begin{abstract}
%\textcolor{red}{HIEU:Correcting pseudo-labels via  bounding box classifier and high-activate indexes similarity.} 
Semi-supervised semantic segmentation methods leverage unlabeled data by pseudo-labeling them. Thus the success of these methods hinges on the reliablility of the pseudo-labels. Existing methods mostly choose high-confidence pixels in an effort to avoid erroneous pseudo-labels. However, high confidence does not guarantee correct pseudo-labels especially in the initial training iterations. In this paper, we propose a novel approach to reliably learn from pseudo-labels. First, we unify the predictions from a trained object detector and a semantic segmentation model to identify reliable pseudo-label pixels. Second, we assign different learning weights to pseudo-labeled pixels to avoid noisy training signals. To determine these weights, we first use the reliable pseudo-label pixels identified from the first step and labeled pixels to construct a prototype for each class. Then, the per-pixel weight is the structural similarity between the pixel and the prototype measured via rank-statistics similarity. This metric is robust to noise, making it better suited for comparing features from unlabeled images, particularly in the initial training phases where wrong pseudo labels are prone to occur. We show that our method can be easily integrated into four semi-supervised semantic segmentation frameworks, and improves them in both Cityscapes and Pascal VOC datasets.  Code is available at \url{https://github.com/cvlab-stonybrook/Weighting-Pseudo-Labels}.

\end{abstract}
\section{Introduction}
\begin{figure}[ht!]
    \centering
   \includegraphics[width=1\linewidth ]{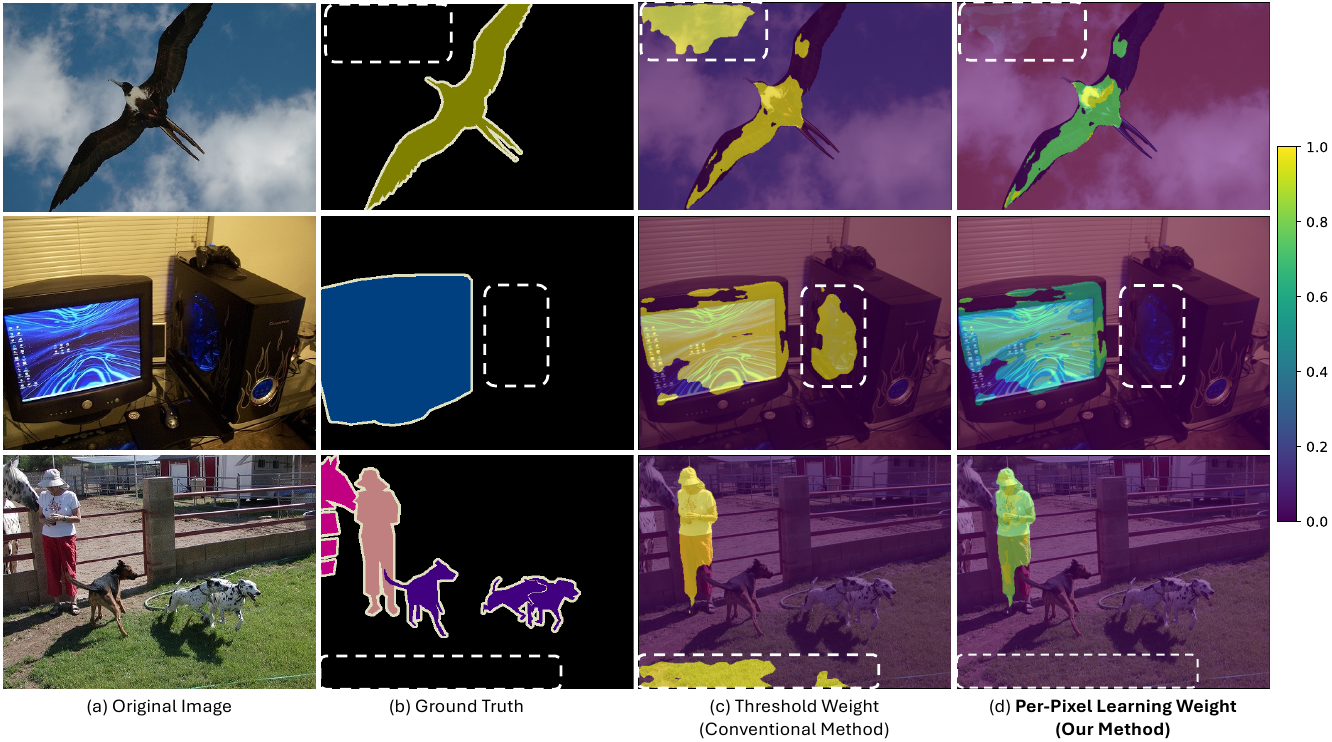}
 
 \vspace{-.1in}
\caption{Per-pixel Learning Weight Visualization (heat-map). Our Per-pixel Learning Weight shows that the weight on unreliable high-confidence pseudo-labels (dotted white box) is reduced in contrast to conventional confidence thresholding ($\ge 0.95$). Pseudo-labels are generated using AugSeg \cite{zhao2023augmentation} after 50 epochs for $\frac{1}{16}$ Pascal VOC Dataset. }
  \label{fig:teaser}
   \vspace{-.1in}
\end{figure}
\vspace{-.1in}
%%%%%%professor
%per pixel confidence  factor correctly identifies the unreliable high confidence pseudo-labels shown in the white box

Semantic segmentation is essential for various applications, including autonomous driving
\cite{badrinarayanan2017segnet,kundu2014joint}, and drone imagery \cite{umar2021forest,kattenborn2019convolutional}.
However, current models require large-scale pixel-level annotations for training, which are laborious and expensive to collect\cite{JingyiICCV21,Xu2022GeneratingRS,Xu2023FSOD,Xu2023ZeroShotOC}. Semi-supervised segmentation \cite{alonso2021semi,hu2021semi} alleviates this data dependency by learning from a limited set of labeled images and numerous unlabeled images.

An effective semi-supervised strategy is pseudo-labeling via a teacher-student framework \cite{sohn2020fixmatch,lee2013pseudo,zou2020pseudoseg,yang2022st++}. 
This strategy typically involves using a teacher model trained on limited labeled images to pseudo-label the unlabeled ones and then using these pseudo-labels as additional supervision to train the student model. 
The correctness of the pseudo labels is a major concern, as wrongly pseudo-labeled pixels might never be corrected throughout the training process, leading to ``confirmation bias'' \cite{liu2021certainty, tarvainen2017mean}. To remedy this, previous works \cite{zhao2023augmentation, alonso2021semi, sohn2020fixmatch, zuo2021self, liu2022perturbed, fan2022ucc, hu2021semi} only pseudo-label pixels with high confidence scores. However, the confidence score is a sub-optimal proxy for  pseudo-label correctness, especially in early training epochs. For example, in the initial training epochs on Pascal VOC dataset ($\frac{1}{16}$ data partition) with a confidence threshold of 0.95, $\sim$20\% of the pseudo-labels are still incorrect (see Section \ref{plnoiseandrankstats}).
% For example, when training on Pascal VOC dataset for $\frac{1}{16}$ data partition, the initial training epochs with a confidence threshold of 0.95, $\sim$20\% of the pseudo-labels are still incorrect (see Sec \ref{plnoiseandrankstats}).

To address this issue, we introduce a novel approach for estimating the reliability of each pixel's pseudo-label and then assign a learning weight to each pixel based on its reliability measure.
The key observation is that pixels belonging to the same category frequently share a subset of k maximally activated representation components, \ie, top-k rank statistics \cite{Yagnik2011ThePO, han2020automatically}. For example, considering the last layer features of a network pre-trained on the Pascal VOC Dataset, we observe that pixel representations in the ``dog'' category typically exhibit the top-5 highest magnitudes in the 10\textsuperscript{th}, 71\textsuperscript{st}, 97\textsuperscript{th}, 98\textsuperscript{th}, and 111\textsuperscript{th} dimensions of their 256-dim feature embeddings. Therefore, any pixel whose pixel representation exhibits the top-5 highest magnitudes in the same feature dimensions (i.e., 10\textsuperscript{th}, 71\textsuperscript{st}, 97\textsuperscript{th}, 98\textsuperscript{th}, and 111\textsuperscript{th} dimensions) has a high probability of belonging to the ``dog'' category. Notably, this pattern of index-based consistency is present even during the early stages of training the segmentation model with limited training data (see Section \ref{plnoiseandrankstats}). On the other hand, the magnitude of each component in the embedding tends to vary significantly during the training, making  value-based metrics such as entropy or confidence-score less reliable. As can be seen in Fig. \ref{fig:teaser}, even pixels with very high confidence scores ($\geq0.95$) can still have incorrect pseudo-labels (Fig. \ref{fig:teaser} (c)). However, these pixels are assigned very small weights by our method (Fig. \ref{fig:teaser} (d)) since their top-k maximally activated representation components are inconsistent with the majority of the same category pixels.

The question is  how to identify these subsets of k maximally activated representation components for each class.  To do so, we construct a class pixel-prototype by using labeled pixel representations and selecting a set of highly reliable pixel representations from unlabeled pixels.  While previous methods simply define reliable pixels as the ones with high confidence scores from the segmentation model, we further improve this by training an additional object detection model and use the trained segmentation and the object detection model as an ensemble model to identify reliable pixels. We assert that if they predict the same label for a pixel, then we consider the pseudo label of the pixel highly reliable. The agreement between these two models indicate the pseudo-label's correctness because they have distinct underlying inductive mechanisms: while the object detector assigns a single label to a group of pixels based on a holistic view of the image crop, the segmentation model assigns a label for each pixel based on the ``\textit{local}'' patch and the surrounding context. In our experiments, this ensemble model identifies these reliable pixels more accurately, compared to a single segmentation model. Note that we only train this additional object detector from scratch on the limited labeled images and only for object-based categories. Reliable pixels for categories such as ``sky'' and  ``building'' are obtained only from labeled pixels.

We incorporate our method into the four semi-supervised segmentation methodologies—UniMatch \cite{yang2023revisiting}, AugSeg \cite{zhao2023augmentation}, AEL \cite{hu2021semi}, and U2PL \cite{wang2022semi}—notably improving segmentation results for each across all data partitions in Pascal VOC\cite{everingham2010pascal} and  Cityscapes \cite{cordts2016cityscapes}  datasets. In summary, our contributions are:

\begin{enumerate}
%\itemindent=10.87pt
%\begin{enumerate}
    \item We propose a novel method for weighing pseudo-labels to alleviate the potential noisy pseudo-label issue in semi-supervised segmentation via comparing top k maximally activated representation components.
    \item We propose a novel method to identify reliable pixels by unifying the predictions from object detection and semi-supervised semantic segmentation models. The object detector is trained solely on the limited labeled data. 
    \item We show that our method can be easily integrated into other approaches by integrating it into four state-of-the-art approaches and getting consistent improvements across all settings. 
\end{enumerate}

\section{Related Work}
Semi-supervised learning (SSL) is a heavily studied problem. Recent works in SSL has been categorized into consistency regularization \cite{bachman2014learning,french2019semi,ouali2020semi,sajjadi2016regularization,yang2022st++,durasov2024enabling,durasov2024zigzag}, entropy minimization \cite{chen2021semisupervised,grandvalet2004semi} and pseudo-labeling \cite{lee2013pseudo,sohn2020fixmatch,Le_CVPRW19,Le_RS22,Le_ICCV2017,Le_ECCV18}. Here, we focus on pseudo-labeling and consistency regularization. 

\noindent
\textbf{Pseudo labeling:}
Pseudo-labeling \cite{lee2013pseudo,shi2018transductive} and self-training \cite{yarowsky1995unsupervised,mcclosky2006effective} aim to train a model on labeled images to generate pseudo-labels for unlabeled data. Because pseudo-labels are noisy, most approaches \cite{zou2020pseudoseg,selvaraju2017grad,sohn2020fixmatch,rizve2021defense} are focused on refining pseudo-labels. For example, PseudoSeg \cite{zou2020pseudoseg} focuses on improving the quality of pseudo-labels using grad-cam \cite{selvaraju2017grad} based attention. 
% FixMatch \cite{sohn2020fixmatch} uses a confidence threshold to select reliable pseudo-labels. UPS \cite{rizve2021defense} uses uncertainty to select reliable pseudo-labels. 
To select reliable pseudo-labels, FixMatch \cite{sohn2020fixmatch} uses confidence thresholding while UPS \cite{rizve2021defense} uses uncertainty.
ST$++$ \cite{yang2022st++}, prioritizes unlabeled images that can provide more reliable pseudo-pixels. 
Recent approaches, AEL\cite{hu2021semi}, propose adaptive frameworks to prefer under-performing categories. U2PL \cite{wang2022semi} considers extracting reliable pseudo-labels from unreliable pseudo-labels. 
CFCG \cite{li2023cfcg} proposed using cross fusion and contour guidance to improve the pseudo-labels. However, all these pseudo-labeling-based approaches rely on a segmentation network trained on limited labeled images to generate pseudo-labels. They further select reliable pseudo-labels based on high confidence (low entropy). Rizve \etal \cite{rizve2021defense} observes that these high confidence predictions on unlabeled images can be incorrect due to poor network calibration \cite{guo2017calibration}. A recent work \cite{ibrahim2020semi} uses ground-truth bounding box annotations in the unlabeled images to improve semi-supervised segmentation. We on the contrary do not use any extra supervision in the unlabeled data. We train our object detector solely on the limited labeled data to generate object proposals in the unlabeled data.

\noindent
\textbf{Consistency Regularization:}
Consistency Learning \cite{sohn2020fixmatch} enforces consistent predictions across different augmentations of unlabeled data. UniMatch \cite{yang2023revisiting} uses perturbation in the feature space to generate different augmentations of unlabeled data. ICT \cite{verma2019interpolation} uses Mixup \cite{zhang2017mixup} augmentation for consistency regularization.  Recently, many methods \cite{french2019semi,yang2022st++,yuan2021simple} use Cutout \cite{devries2017improved}, Cutmix \cite{yun2019cutmix}, Classmix \cite{olsson2021classmix} as strong data augmentation. Consistency Regularization is used with pseudo-labeling techniques in Mixmatch \cite{berthelot2019mixmatch} and TC-SSL \cite{zhou2020time}. In our work, we use a set of weak and strong data augmentations \cite{alonso2021semi,olsson2021classmix, yang2023revisiting} to generate pseudo-labels.
\begin{figure*}[ht!]
         \centering
           \includegraphics[width=0.75\linewidth]{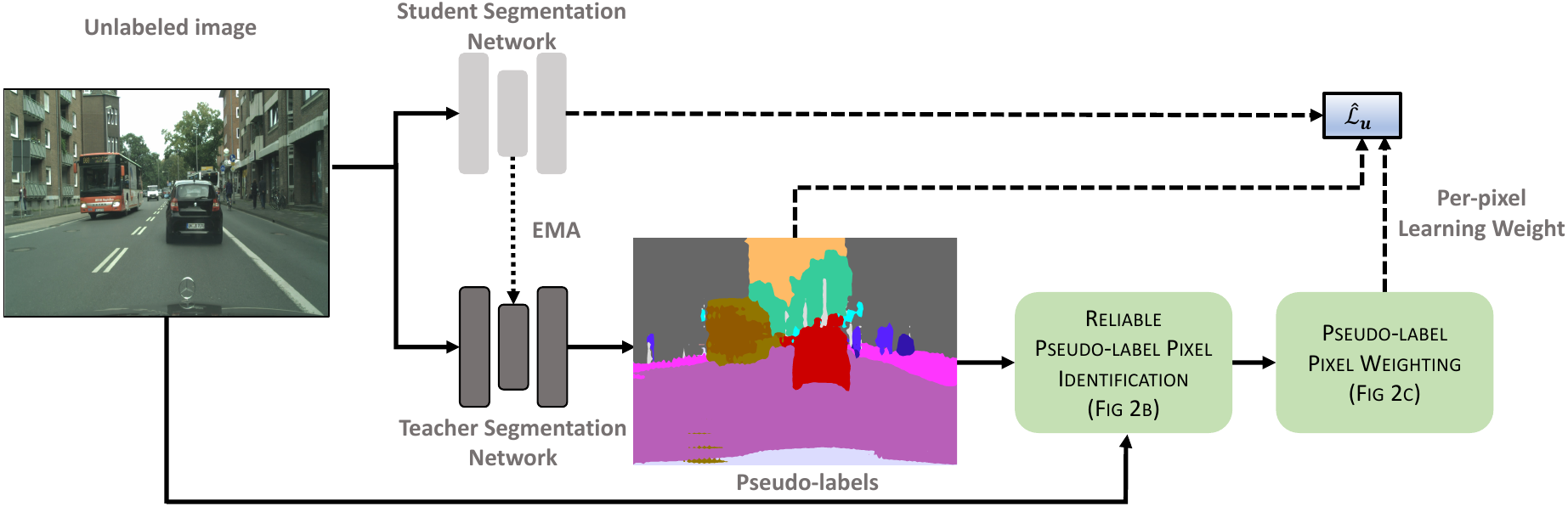} 
           \makebox[\linewidth]{(a) Overall Training Pipeline}\\
           %\vspace{0.5cm}
         %\includegraphics[width=0.9\linewidth]{pics/hieucvprmodel2ndpic.pdf}
         \includegraphics[width=0.9\linewidth]{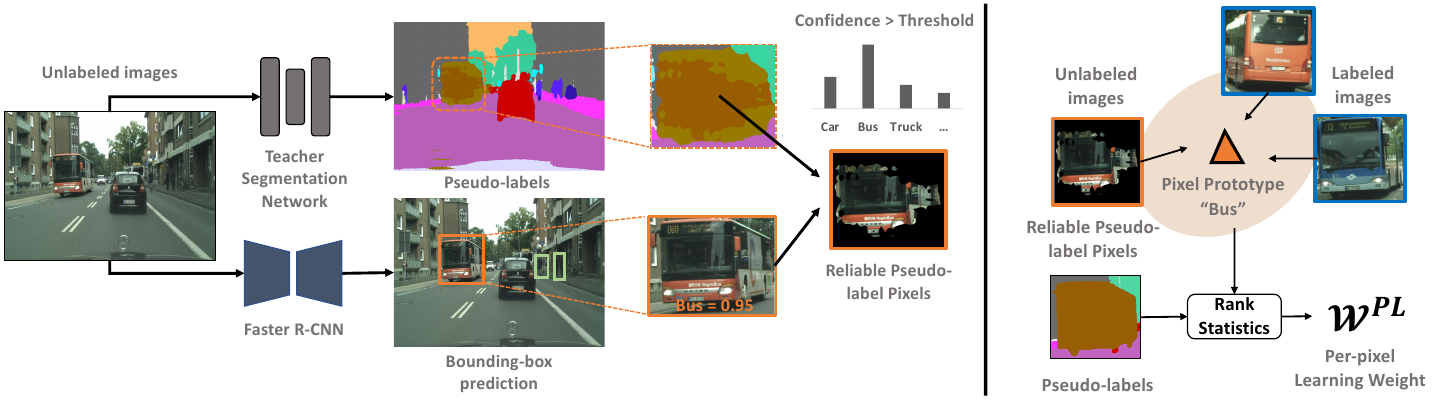}
           \makebox[0.43\linewidth]{(b) Reliable Pseudo-label Pixel Identification  }
\makebox[0.28\linewidth]{\hspace{2.3cm}(c) Pseudo-label Pixel Weighting}
%0.32 \makebox[0.45\linewidth]{(b) Prototype-based Pseudo Labelling}
         % \caption{\textbf{Overall pipeline of our method.} We propose a novel method for pseudo-labeling images for semantic segmentation. (a) The overall training pipeline of our method in Teacher-Student Pipeline (b) We first identify pixels with reliable pseudo-labels using an object detector and entropy-based adaptive thresholds. The reliable pixels are defined as ones being labeled as the same class by both the detection and segmentation model with high confidence scores and with low entropy. (c) We construct a pixel-representation prototype for each class using labeled images and identified reliable pixels. The prototype is built based on the feature representation of labeled pixels and reliable pseudo-labeled pixels of the class ( The prototypes are updated after every iteration ). We then use rank statistics \cite{han2020automatically} to find pixels that likely belong to the same category as the prototype and assign the same label to them.
         % }
         \caption{\textbf{Overall Pipeline of our novel pseudo-labeling based semantic segmentation:} (a) End-to-end Teacher-Student Pipeline (b) We first identify pixels with reliable pseudo-labels using an object detector and segmentation model. The reliable pseudo-label pixels are defined as ones being labeled as the same class by both the detection and segmentation models with high confidence scores. (c) We constructed a pixel-representation prototype for each class using labeled images and identified reliable pseudo-label pixels. We then use rank statistics \cite{han2020automatically} to weight the pseudo-labels predicted by the teacher network.}
        \label{fig:pipeline}
\end{figure*}
\section{Proposed Method}

In semi-supervised semantic segmentation, we are given two distinct data sources: a set of labeled images denoted as $\mD^{l} = \{x^{l}_{i},y^{l}_{i}\}_{i=1}^{N_{l}}$ and a collection of unlabeled images denoted as  $\mD^{u}=\{x^{u}_{i}\}_{i=1}^{N_{u}}$, where $|\mD_{u}| \gg |\mD_{l}|$. The central aim is to develop a semantic segmentation model that utilizes the knowledge from labeled and unlabeled data. This section first discusses the basic framework of semi-supervised semantic segmentation in Preliminary (Section \ref{preliminary}). Next, we provide an overview of our proposed method (Section \ref{overview}). Then we introduce the two steps that underpin our proposed novel method: (1) Reliable Pseudo-label Pixel Identification (Section \ref{Reliable Pseudo-label Pixel Identification}), (2) Pseudo-label Pixel Weighting (Section \ref{Pseudo-label Pixel Weighing}).

%$\hat{\gamma} = \{\hat{y}^u\}$
\subsection{Preliminary}\label{preliminary}
 The conventional framework of semi-supervised semantic segmentation\cite{alonso2021semi, zhao2023augmentation, hu2021semi} is built on top of a student/teacher model. The teacher model shares the same architecture with the student model but uses a different set of parameters, which are updated by the exponential moving average (EMA) of the student model \cite{hu2021semi}. The teacher model generates a set of pseudo labels $\hat{y}^u$ on the weakly augmented unlabeled data $\mD^{u}$. Subsequently, the student model is trained on both  weakly augmented labeled data $\mD^l$ with the ground truth and strongly augmented unlabeled data $\mD^u$ with the generated pseudo labels $\hat{y}^u$. The overall loss consists of the supervised loss $\mL_s$ and the unsupervised loss $\mL_u$:
% \begin{align}
%     %\begin{split}
%     \mL_{s} &= \frac{1}{N_l} \sum_{i=1}^{N_{l}} \frac{1}{WH} \sum_{j=1}^{WH} l_{ce}(y^{l}_{ij},p(x^{l}_{ij})) \label{eq1} \\ 
% %\end{align}
% %\begin{align}
%     %\begin{gathered}
%     \mL_{u} &= \frac{1}{N_u} \sum_{i=1}^{N_{u}}\frac{1}{WH} \sum_{j=1}^{WH} \mathbbm{1}(\textrm{max}(p( \mA^{w}(x^{u}_{ij}) )) \ge \tau)l_{ij}^u \label{eq2} \\
%     l_{ij}^u &= l_{ce}(\hat{y}_{ij}^{u},p( \mA^{s}(\mA^{w}(x^{u}_{ij})) ))\label{eq3}
%     %\end{gathered}
% \end{align}
\begin{align}
    %\begin{split}
    \mL_{s} &= \frac{1}{|\mD^l|} \sum_{x^l \in \mD^l}\frac{1}{WH} \sum_{i=1}^{WH}  l_{ce}(y^{l}_{i},p(x^{w,l}_{i})) \label{eq1} \\ 
%\end{align}
%\begin{align}
    %\begin{gathered}
    \mL_{u} &= \frac{1}{|\mD_u|} \sum_{x^u \in \mD_u} \frac{1}{WH} \sum_{i=1}^{WH} \mathbbm{1}(\textrm{max}(p( x^{w,u}_i )) \ge \tau)l_{i}^u \label{eq2} \\
    l_{i}^u &= l_{ce}(\hat{y}_{i}^{u},p( x^{s,u}_i )  )\label{eq3}
    %\end{gathered}
\end{align}
where $x^{w,l}$ and $x^{w,u}$ are weak augmentations of $x^l$ and $x^u$ respectively, $x^{s,u}$ is strong augmentation of $x^u$. $W$ and $H$ correspond to the width and height of the input image, $l_{ce}$ denotes the standard pixel-wise cross-entropy loss, $p(\cdot)$ is the network prediction for labeled and unlabeled images, $C$ is the number of classes.  $\tau$ is a predefined threshold to filter noisy labels. Further $\hat{y}^{u}_{i} = argmax(p(x^{w,u}_i))$ corresponds to the teachers prediction under weak augmentation view.

Consequently, the overall loss function can be defined as:
\begin{equation}\label{eq4}
 \mL = \mL_{s}+\alpha\mL_{u}   
\end{equation}
where $\alpha$ controls the contribution of the unsupervised loss.

\subsection{Overview}\label{overview}
Fig. \ref{fig:pipeline}(a) presents a comprehensive view of our teacher-student framework method. There are two main ideas: we employ an additional object detector for better identifying reliable pseudo-label pixels, and we use rank statistics with class prototypes to assign a per-pixel learning weight for each pseudo-labeled pixel. 

First, we use an ensemble model comprising of the teacher model and an additional Faster-RCNN \cite{ren2015faster} model to identify reliable pseudo-label pixels (Fig. \ref{fig:pipeline}(b)). We assert that if both models predict the same label for a pixel, then we consider the pseudo label highly certain. We use these reliable pseudo-label pixels and the pixels from the labeled images to construct class prototypes. The class prototypes are used to compute a per-pixel learning weight via rank statistics for training the network - Fig. \ref{fig:pipeline}(c).

\subsection{Reliable Pseudo-label Pixel Identification}\label{Reliable Pseudo-label Pixel Identification}
Semi-supervised segmentation aims to identify reliable pseudo-labels from the unlabeled images and use them for training. Existing methods \cite{sohn2020fixmatch,zou2020pseudoseg,yang2022st++,rizve2021defense,zhu2020improving,yang2023revisiting} pseudo-label unlabeled images after filtering the pseudo-labels predicted by the teacher segmentation network based on confidence or entropy thresholds.

However, these reliable pseudo labels used during the training process are noisy, especially in the initial stages of training, due to the limited labeled data for training and poor model calibration \cite{rizve2021defense}. To overcome this challenge, we propose a solution to reduce the reliance on segmentation model confidence by using an ensemble of teacher and an additional object detection model (Faster R-CNN \cite{ren2015faster}).
Specifically, we first train a Faster R-CNN \cite{ren2015faster} only on the limited labeled data $\mD^l$. We use the generated object proposals to complement the segmentation network in identifying a set of reliable pseudo-label pixels.
We use two criteria to determine if a pixel is a reliable pseudo-label:

\begin{enumerate}
    \itemindent=10.87pt
    \item The segmentation and the detection model label the pixel as the same class.
    \item The pixel confidence is higher than a predefined threshold.
\end{enumerate}
The first criterion ensures that we exclusively consider pixels within the object proposal that share the same class as predicted by the segmentation network, i.e., only ``\textit{bus}'' pixels in Fig. \ref{fig:pipeline} (b). The second criterion considers pixels within the object proposal, on which the segmentation network is most confident, similar to previous work \cite{zhang2022mfnet,guan2022unbiased, alonso2021semi,yang2023revisiting}. Applying these two criteria, we identify a set of reliable pseudo-labeled pixels from the unlabeled images. The key point here is that if the two models agree over the label of a pixel, we consider the pseudo-label highly reliable because they have distinct underlying inductive mechanisms. While the object detector assigns a single label to a group of pixels based on the holistic view of the image crop, the segmentation model assigns a label for each pixel based on the ``\textit{local}” patch and the surrounding context.

Note that we only re-purpose the segmentation labels in $D_l$ to train the Faster R-CNN model. Given an image and its semantic segmentation mask, for each category, we define an object box for each set of connected pixels as the smallest bounding box containing them. Each box might contain more than one object, but it is not an issue for our semi-supervised segmentation method. We discard bounding boxes of background classes such as sky, vegetation, or buildings since their bounding boxes tend to cover almost the entire image. As we do not train the object detector on those background classes, reliable pixels for those classes only come from labeled images.
\begin{figure*}[ht!]
         \centering
           \includegraphics[width=0.85\textwidth]{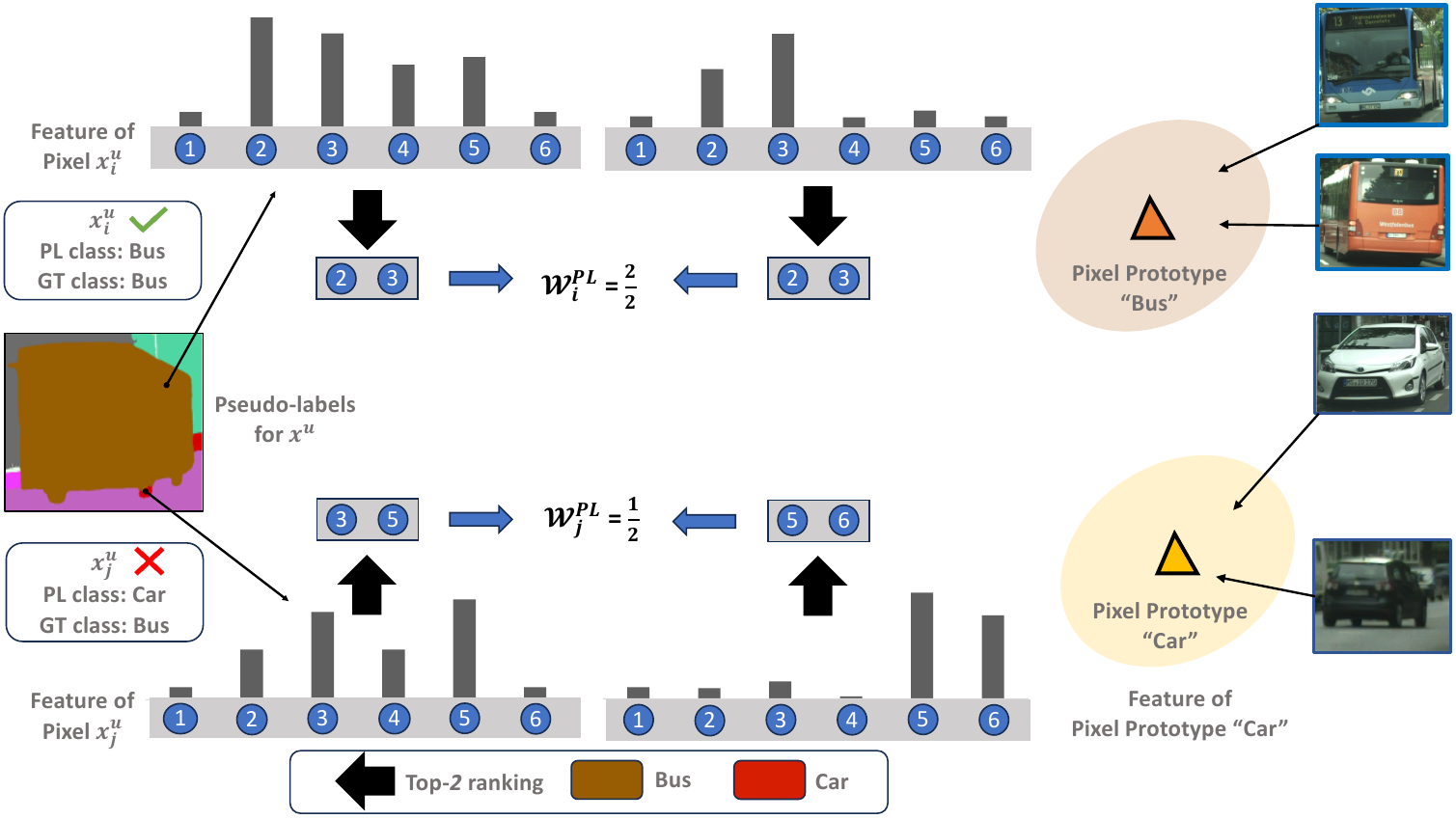}
           %\makebox[\linewidth]{(a) Pseudo-label Pixel Weighting}\\

         \caption{\textbf{Demonstration of Pseudo-label Pixel Weighting via rank-statistics:} This diagram shows the top-$2$ ranking based pseudo-pixel weighing for two pixels $x^u_i$ and $x^u_j$ in unlabeled mage $x^u$. PL class is Pseudo-label class, GT class is Ground Truth class. \textbf{Note, top-2 ranking is same between $x^u_i$ and bus pixel prototype, while different between $x^u_j$ and car pixel prototype}. } 
        \label{fig:plrankstats}
\end{figure*}

\subsection{Pseudo-label Pixel Weighting}\label{Pseudo-label Pixel Weighing}
We assign an adaptive per-pixel learning weight to each pseudo-labeled pixel to avoid noisy training signals. To determine this weight, we first use the identified reliable pseudo-label (Section \ref{Reliable Pseudo-label Pixel Identification}) and labeled pixels to construct a prototype for each class. Then, the per-pixel weight is the structural similarity between the pixel and the prototype measured via rank-statistics~\cite{Yagnik2011ThePO, han2020automatically}. This metric only considers the overlap between indices of the highest value elements (top-$K$) of the two representation vectors, \textit{i.e.}, which feature components are activated the most. Specifically, given two features $z_i$ and $z_j$, we rank the feature dimensions in the vector $z_i$ and $z_j$ by their magnitude. We consider two features belonging to the same class if the set of their top-k ranking of their feature dimensions match $\{top_k(z_i) = top_k(z_j)\}$. Rank statistics similarity is not based on raw feature magnitudes but rather the structural similarity between vector representations, making rank statistics similarity more robust to noise \cite{Yagnik2011ThePO, maturi2008new, pinto2005weighted, shieh1998weighted} when comparing high-dimensional feature representations. Thus, it is better suited for comparing features in the initial training stages of semi-supervised segmentation when the model is poorly calibrated, and the features and predictions are noisy \cite{rizve2021defense} (See Section \ref{plnoiseandrankstats} for analysis on rank statistics being more robust to noise). Here, we use rank statistics for features extracted from the second-to-last layer of the segmentation model \cite{alonso2021semi}. It serves as a secondary source of confirmation besides the confidence score to verify the correctness of the pseudo-labels. 

Specifically, we first construct a per-class feature prototype using pixel features in labeled images and reliable pseudo-label pixels in unlabeled images (Section \ref{Reliable Pseudo-label Pixel Identification}). The prototype is the mean of the pixels' latent embeddings, which are the outputs of the penultimate layer of the segmentation model at the pixel locations. It is computationally expensive to extract features from all labeled and unlabeled images in each iteration. So, we save their features in a feature memory bank \cite{alonso2021semi},  which we query to generate the per-class feature prototype (more details on memory bank in supplementary material for details).
%Inspired by the observation in the context of ranking, that agreement among high-ranking coefficients is more important than the rest \cite{pinto2005weighted,shieh1998weighted}.  We rank the values in feature vectors by magnitude and compare the similarity in the sets of the top-k ranked dimensions between the pixel and the prototype feature.

For each unlabeled image $x^u$, we compute a per-pixel learning weight $\mW^{PL} \in \mathbbm{R}^{W \times H}$, where $W$ and $H$ are the width and height of the image. We use a soft extension of ranking statistics \cite{han2020automatically} proposed in \cite{zhao2021novel} as our similarity metric. It measures the similarity between two features as the number of shared elements in their sets of top-k ranking. 
%Motivated by Zhao \etal\cite{zhao2021novel}, which proposes a soft extension of  ranking statistics \cite{han2020automatically}, where the 
 The per-pixel learning weight $\mW^{PL}$ of the pseudo-label at the position $i$ in an unlabeled image $x^u$, which has been assigned a pseudo-label of class $c$ by the teacher model, is defined as $\mW^{PL}_{i} = \frac{s}{k}$. Here, $s$ represents the count of common elements within the sets of top-$k$ ranking of the pixel feature at position $j$  and the feature prototype for class $c$ (we use $k=5$ for all our experiments). We illustrate in Fig. \ref{fig:plrankstats} that per-pixel weighting based on our approach for top-$2$ ranking based feature similarity. We observe that pseudo-labels misclassified as ``car'' are provided lower weights than  pseudo-labels correctly classified as ``Bus''.
 
% We define the weight $w_{ij}$ of the pseudo-label at position $i,j$ of an unlabeled image $x^u$, which has been pseudo-labeled by the teacher as class $c$, as $w_{ij} = \frac{c}{k} \in [0,1]$, where $c$ is the number of shared elements in the top-k ranked dimensions $top_{k}(z_{ij})$ and $top_{k}(z^{c})$. $z_{ij}$ and $z^{c}$ are the features of the  pixel at position $i,j$, and feature prototype of class $c$.
We modify the unsupervised loss of the conventional teacher-student framework (Equation \ref{eq2}) to incorporate the per-pixel learning weight $\mW^{PL}$. 
Hence, the unsupervised loss  for our approach is:
\begin{align}\label{eq4}
    %\begin{gathered}
    \mL_{u} &= \frac{1}{|\mD_u|} \sum_{x^u \in \mD_u} \frac{1}{WH} \sum_{i=1}^{WH} \mathbbm{1}(max(p( x^{w,u}_i )) \ge \tau)l_{i}^u \\
    l_{i}^u &= \mW^{PL}_{i}l_{ce}(\hat{y}_{i}^{u},p( x^{s,u}_i ))
    %\end{gathered}
\end{align}
% In Equation \ref{eq4}, we observe that $\mW^{PL}$ reduces the weight of potential high confidence noisy pseudo-labels in the training of the student.
Thus, our overall loss function  consists of the supervised loss (Equation \ref{eq1}) and the unsupervised loss (Equation \ref{eq4}), is:
\begin{equation} \label{eq5}
    \begin{split}
    \mL = \mL_{s}+\alpha \mL_{u}
   \end{split}
\end{equation}
where $\alpha$ controls the amount of contribution of the unsupervised loss.

\section{Experiments}
\subsection{Setup}
\noindent
\textbf{Datasets:}  The \textbf{PASCAL VOC 2012} dataset \cite{everingham2010pascal} is a widely recognized benchmark for semantic segmentation, with 20 object categories and an additional background class. It is partitioned into training, validation, and testing subsets, containing $1464$, $1449$, and $1556$ images (\textit{Classic}), respectively. Following \cite{zhao2023augmentation, chen2021semi}, we also include the additional augmented dataset \cite{hariharan2011semantic} (\textit{Blender}), which includes a collection of $10582$ training images. We adopt the same partition protocols in \cite{zhao2023augmentation, chen2021semi} to evaluate our method in both \textit{Classic} and \textit{Blender} sets. The \textbf{Cityscapes} dataset \cite{cordts2016cityscapes}, tailored for urban scene analysis, comprises of 30 classes, though only 19 of these are employed for scene parsing assessments. This dataset is divided into $2975$ training images, $500$ validation images, and $1525$ testing images.

\noindent
\textbf{Implementation Details:} For fair comparison, we adopt DeepLabv3+ \cite{chen2018encoder} as the decoder in all of our experimental setups, and compare with both ResNet-101 and ResNet-50 \cite{he2016deep} as the backbone architecture. We incorporate our method into the framework of four semi-supervised segmentation methods:  AugSeg \cite{zhao2023augmentation}, AEL \cite{hu2021semi} ,U2PL \cite{wang2022semi} and Unimatch \cite{yang2023revisiting}. Our method serves as a pseudo-label weighting strategy to alleviate the influence of confident, noisy pseudo-labels during training, without changing their original architecture and training procedures.
Consistent with common practices \cite{wang2022semi}, we train our models on the Cityscapes and Pascal datasets at resolutions of 801 and 513, respectively. It is important to note that in the interest of maintaining a fair comparison, our approach, labeled as ``UniMatch+Ours'' employs a training resolution of 321, aligning with the resolution used by UniMatch for training on the Classic set of the Pascal VOC dataset. Further when training a baseline integrated with our method, we use the same weak and strong augmentations as used by the corresponding baseline.

After the Faster R-CNN is trained on only the limited labeled dataset, the confidence threshold to select bounding boxes in unlabeled images is set at $0.95$ for Cityscapes and $0.85$ for the Pascal Dataset, respectively. In all our experiments  we set $K=5$ for rank statistics.

\subsection{Comparison with State-of-the-Art Methods} 
We conduct experiments on two popular benchmarks: PASCAL VOC 2012 and Cityscapes. We integrate our method to four  semi-supervised methods: AugSeg \cite{zhao2023augmentation}, AEL \cite{hu2021semi} ,U2PL \cite{wang2022semi} and Unimatch \cite{yang2023revisiting}. Note, UniMatch \cite{yang2023revisiting} is a consistency regularization based method. The results demonstrate
consistent improvement in performance over the corresponding baselines across all partitions. This notable improvement across datasets robustly validates the effectiveness of our proposed approach.\\
\textbf{PASCAL VOC 2012 Dataset.} Table \ref{tab:voc} presents a comparative analysis with other SOTA methods for both the \textit{Classic} and \textit{Blender} sets.  Our method consistently enhances the performance of all baseline methods across all data partitions for both the \textit{Classic} and \textit{Blender} sets. In particular, the most significant improvements are observed in the partition with the least labeled data ($\frac{1}{16}$), where our method boosts the performance of AugSeg \cite{zhao2023augmentation}, AEL \cite{hu2021semi} ,U2PL \cite{wang2022semi} and Unimatch \cite{yang2023revisiting} by $2.0\%$, $3.7\%$,$3.1\%$ and $1.5\%$ respectively on the \textit{Classic} set and $1.9\%$, $3.3\%$,$2.9\%$ and $2.1\%$ respectively on the \textit{Blender} set. \\
\textbf{Cityscapes Dataset.} Table \ref{tab:ct} presents a comparative analysis with other SOTA methods. Our method consistently enhances the performance of all baseline methods across all data partitions. We observe that similar to results in PASCAL VOC 2012 dataset, our method brings the biggest improvement for the least labeled data partition ($\frac{1}{16}$), improving performance of AugSeg \cite{zhao2023augmentation}, AEL \cite{hu2021semi} ,U2PL \cite{wang2022semi} and Unimatch \cite{yang2023revisiting} by $2.1\%$, $3.4\%$,$3.1\%$ and $1.8\%$ respectively for ResNet-101 based encoder.

The consistent improvement in semi-supervised segmentation performance in both datasets and its ability to integrate in different methods substantiates the importance of our method  for segmentation in the limited data domain.

\input{table/voc}
\input{table/ct}

%-----------------------------------------------------------------------------------------------
\begin{figure*}[ht!]
         \centering
           \includegraphics[width=0.97\textwidth]{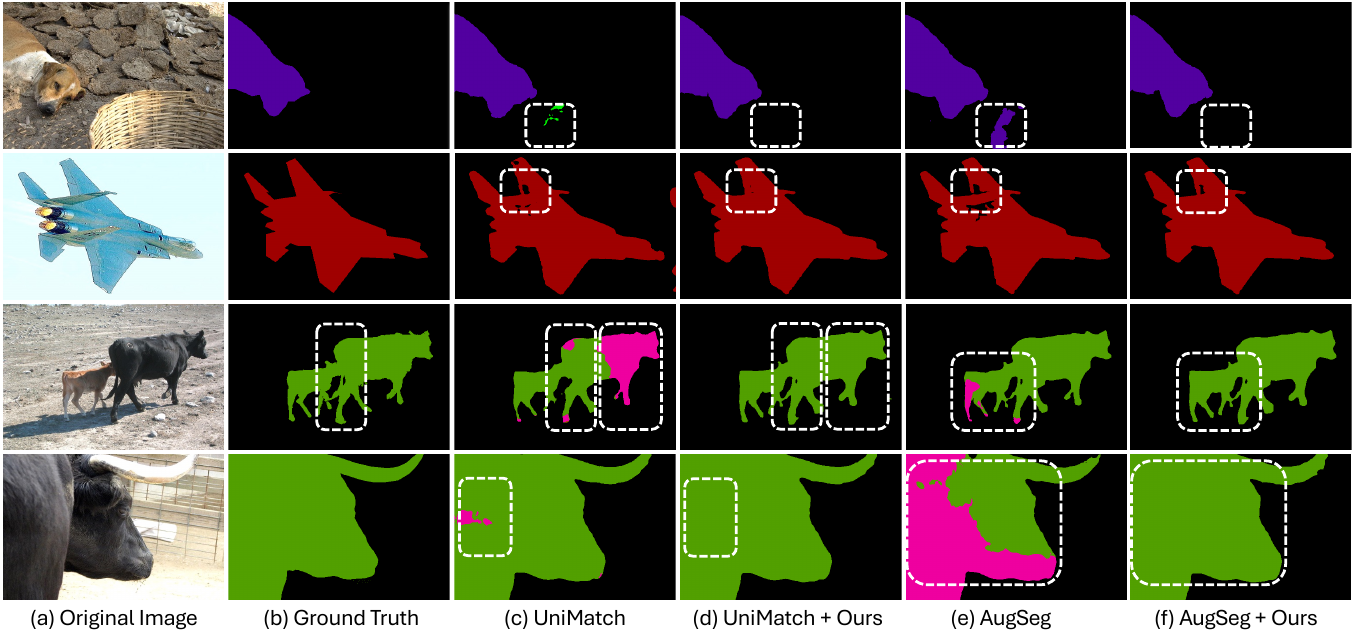} 

         % }
         \caption{\textbf{Qualitative Results on Pascal VOC} From left to right: original image, ground truth, UniMatch   \cite{yang2023revisiting}, UniMatch \cite{yang2023revisiting} + Ours, AugSeg \cite{zhao2023augmentation}, AugSeg \cite{zhao2023augmentation} + Ours. The dotted white box shows the regions where our method improves the baseline.   } 
        \label{fig:allsegmentation}
\end{figure*}
\noindent
\textbf{Qualitative Results}
Figure \ref{fig:allsegmentation} shows the results of different methods evaluated on the Pascal Validation set. Our method incorporated into both UniMatch \cite{yang2023revisiting} and AugSeg \cite{zhao2023augmentation} shows clear improvements over their baselines. Both baselines get benefited from per-pixel pseudo label weighting, helping them achieve better segmentation performance. All methods are trained on $\frac{1}{16}$ data partition.

\section{Ablation Studies}
\begin{figure}[ht!]
    \centering
   \includegraphics[width=1.01\linewidth ]{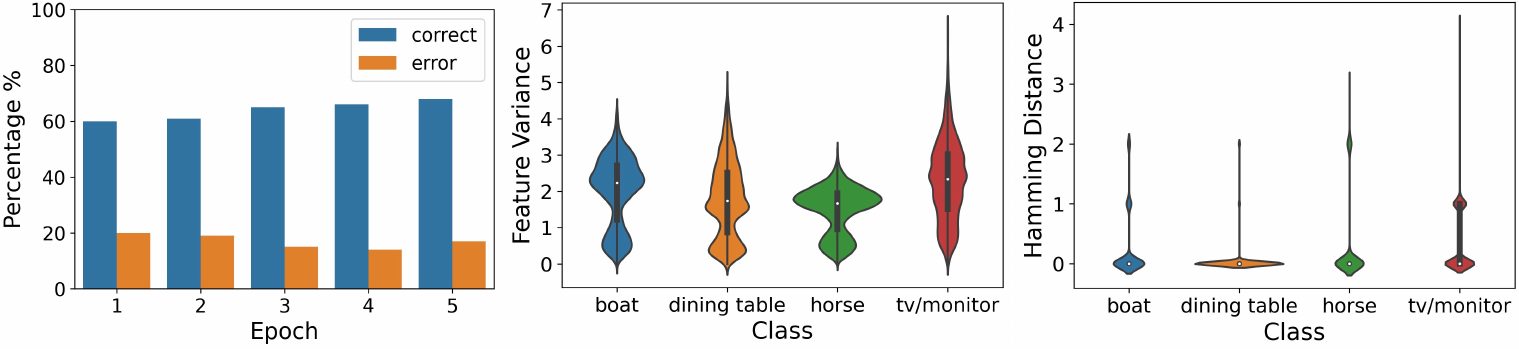}
 
 %\vspace{-.1in}
\caption{(a) The distribution of correct and incorrect pseudo-labels on Pascal VOC dataset above the confidence threshold of $0.95$.  (b) the variation of the features of correct pseudo-labels of 4 random classes, (c) the hamming distance between the binary embeddings of class prototypes of the most confident correct predictions in labeled images and the features of correct pseudo-labels. 
 }
  \label{fig:percentageerror}
   %\vspace{-.1in}
\end{figure}

We conduct extensive experiments to study the impact of various components of our approach. Our experiments were conducted using the Pascal VOC dataset (\textit{Classic}), focusing on one of its most challenging data partitions, $\frac{1}{16}$. For each of these studies in this section, we have selected UniMatch \cite{yang2023revisiting} as the baseline. \\
\noindent
\textbf{Analysis of pseudo-label noise and ranking based feature similarity}\label{plnoiseandrankstats}
\textit{The crux of our approach is we rely on the ranking of feature dimensions based on their magnitudes rather than their raw magnitudes in finding reliable pseudo-labels}.  Here we analyse high confidence noisy pseudo-labels during training and the rationale behind using ranking of feature dimensions as similarity measure. Conventional pseudo-labeling approaches remove noisy pseudo-labels by confidence based thresholding. We observe in Fig. \ref{fig:percentageerror} (a)
that even \textbf{high confidence ($\mathbf{\ge 0.95}$ ) pseudo-labels have significant proportions of incorrect pseudo-labels}. This observation validates that in the initial training epochs, the teacher network is miscalibrated leading to noisy pseudo-labels having high confidence \cite{rizve2021defense}. 

Further, we analyse the features of the classes to understand why feature dimension based ranking is a good similarity metric.
We first compare the variance of the features of correct pseudo-labels of $4$ random classes in Fig. \ref{fig:percentageerror} (b). Further, we generated  feature prototypes based on the most confident correct predictions (confidence threshold $\ge 0.95$) of these classes in the labeled images. From the the class feature prototypes and the correct pseudo-labels we generate binary embeddings with the same dimension as the original features. With the indexes set to $1$ based on the indexes of the original features with top-5 highest magnitudes. We calculate per-class hamming distance between the corresponding binary embeddings of the prototype and the correct pseudo-labels, as illustrated in Fig. \ref{fig:percentageerror} (c). Based on results in Fig. \ref{fig:percentageerror} (b) and (c), \textbf{we observe that feature values show more variation compared to indexes of the top-5 indexes based on magnitude}. This observation that rank ordering of the feature dimensions have less variation than their magniitudes aligns with previous works\cite{Yagnik2011ThePO, maturi2008new, pinto2005weighted, shieh1998weighted}.\\

\begin{figure}[hbt!]
\includegraphics[width=0.8\columnwidth]{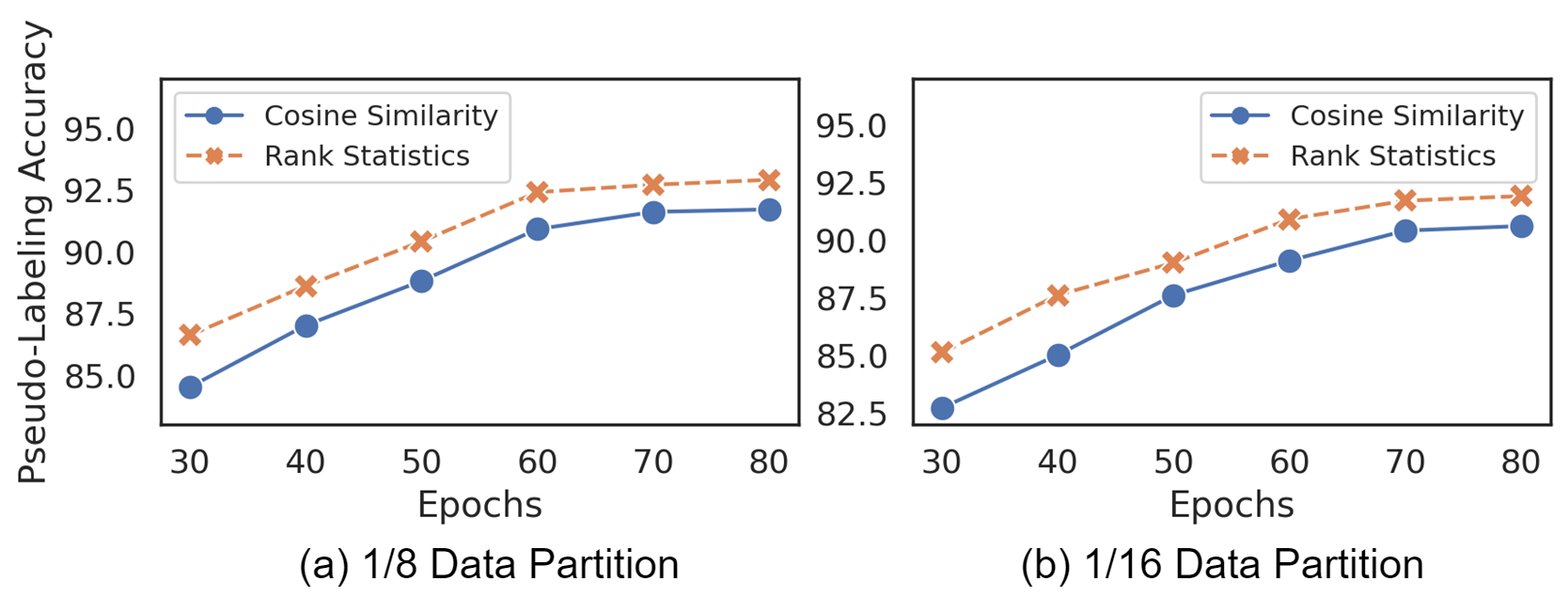}
\centering
\caption{Comparison of pseudo-labeling accuracy in the Pascal VOC unlabeled dataset, between two approaches of measuring similarity  to assign the per-pixel Learning Weight $\mW^{PL}$ (Section \ref{Pseudo-label Pixel Weighing}).}
\label{fig:pseudovscosine}
\end{figure}
\begin{figure}
    \centering
    \begin{minipage}[b]{.47\linewidth}
        \centering
        \includegraphics[width=0.9\linewidth]{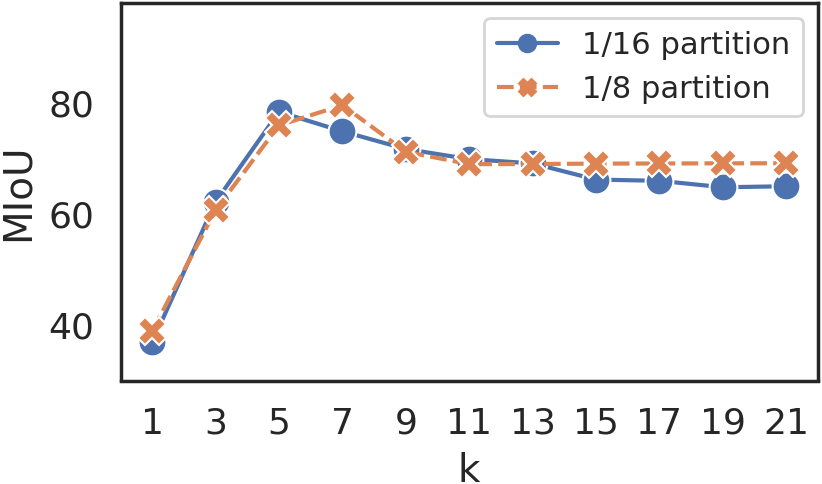}
        \captionof{figure}{Performance evolution with respect to k, for 1/16 and 1/8 partition protocols    Pascal VOC Dataset}
        \label{fig:ablation_k_evolution}
    \end{minipage}\hfill
    \begin{minipage}[b]{.47\linewidth}
    \centering
    \begin{tabular}{c|c|ccl}
    \toprule
    \textbf{RPPI} & \textbf{PPW} & \begin{tabular}[c]{@{}c@{}}1/16\\ (92)\end{tabular} & \begin{tabular}[c]{@{}c@{}}1/8\\ (183)\end{tabular} & \begin{tabular}[c]{@{}c@{}}1/4\\ (366)\end{tabular} \\ \midrule
        &     & 63.1          & 67.4          & 70.8          \\
        & \checkmark   & 67.0          & 70.1          & 73.3          \\
    \checkmark   & \checkmark   & \textbf{70.5} & \textbf{73.3} & \textbf{75.1} \\ \bottomrule
    \end{tabular}
    \captionof{table}{Ablation study of different components: Reliable Pseudo-label Pixel Identification (RPPI) and  Pseudo-label Pixel Weighting (PPW). RPPI and PPW both improve the performance}    
    \label{tab:componentabl}
    %\vspace{2em}
    \end{minipage}
\end{figure}

\noindent
\textbf{Comparison with cosine similarity}
% Cosine-Similarly vs Rank-Statistics based Similarly
Our approach uses rank statistics to calculate the similarity between pixel features at a pseudo-pixel with a class prototype to assign per-pixel leaning weights (Section \ref{Pseudo-label Pixel Weighing}). A conventional approach is  to use cosine-similarity. We compare the two approaches based on pseudo-label accuracy on both $\frac{1}{16}$ and $\frac{1}{8}$ partitions of the Pascal VOC dataset (Fig. \ref{fig:pseudovscosine}). This confirms that rank statistics-based similarity performs better in  comparing noisy features \cite{Yagnik2011ThePO}. A plausible reason is, unlike cosine similarity, rank statistics matches features that share the same feature index ranking rather than their magnitude.\\
\noindent
\textbf{The Effectiveness of Different Components of Our Approach}
We ablate each component of our method step by step. Table \ref{tab:componentabl} reports the studies. We use the basic teacher-student framework in Section \ref{preliminary} as our baseline, which achieves MIoU of $63.1$, $67.4$ and $70.18$ under $\frac{1}{16}$, $\frac{1}{8}$ and $\frac{1}{4}$ partition protocols respectively. As shown in the table, Pseudo-label Pixel Weighting (PPW) introduces a prototype-based pseudo-label per-pixel learning weight achieving an improvement of $3.9\%$, $2.7\%$ and $2.5\%$ under $\frac{1}{16}$, $\frac{1}{8}$ and $\frac{1}{4}$ partition protocols respectively (class prototypes are only from labeled pixels). Reliable Pseudo-label Pixel Identification (RPPI) and and PPW together boost the performance by $7.4\%$, $5.9\%$, and $4.3\%$, demonstrating the effectiveness of our method (class prototypes from labeled pixels and reliable pseudo-labels).\\
\noindent
\textbf{Impact of $k$ in top-k rank statistics}
We analyze how our method performs with respect to $k$. The results on the 1/16 and 1/8 partition protocols of the Pascal VOC dataset are in Fig. \ref{fig:ablation_k_evolution}. We observe that 
$k=\{5,7\}$ gave the best results, further low values of $k$ lead to lower semi-supervised segmentation performance. A potential reason is that the number of dimensions to match is too few thus multiple prototypes can share the same top-$k$ dimensions, leading to pixels being matched with the wrong prototypes. A large value of $k$, however, makes it harder to match pixels with a prototype.  This observation  validates that  in the context of ranking, that agreement among high-ranking coefficients is more important than the rest \cite{pinto2005weighted,shieh1998weighted}. For all our experiments we use $k= 5$.\\
% \begin{figure}[ht!]
%     \centering
%    \includegraphics[width=0.5\linewidth]{pics/MIoU_vs_k.png} 
% \caption{Performance evolution with respect to k, for 1/16 and 1/8 partition protocols of Pascal VOC Dataset
%  }
%   \label{fig:ablation_k_evolution}
% \end{figure}
\textbf{Per-class performance of our method}
We compare the per-class performance of our method incorporated into UniMatch \cite{yang2023revisiting} with respect to the baseline under $\frac{1}{16}$ Pascal VOC (\textit{Classic}) partition. Table \ref{tab:perclassperformance} shows the results. We observe that our method improves the baseline across all classes. Using our rank statistics based pseudo-label weighting approach improves the performance of confusing classes like \textit{Sheep} ($\mathbf{3.7\%}$) and \textit{Sofa} ($\mathbf{3.4\%}$). These classes are often confused with Dog and Chair respectively.

\begin{table*}[hbt!]
\caption{Per-class performance of our method incorporated into UniMatch \cite{yang2023revisiting}, with respect to the baseline. Both methods are trained on $\frac{1}{16}$ Pascal (\textit{Classic}) set. }
\label{tab:perclassperformance}
\begin{subtable}{1.0\linewidth}
\centering
\resizebox{0.98\linewidth}{!}{
\begin{tabular}{@{}c|c|ccccccccccccccccccccc@{}}
\toprule
Method    & \rotatebox{90}{mIoU}              & \rotatebox{90}{Background} & \rotatebox{90}{Aeroplane} & \rotatebox{90}{Bicycle} & \rotatebox{90}{Bird} & \rotatebox{90}{Boat} & \rotatebox{90}{Bottle}  & \rotatebox{90}{Bus} & \rotatebox{90}{Car} & \rotatebox{90}{Cat} & \rotatebox{90}{Chair} & \rotatebox{90}{Cow} & \rotatebox{90}{DiningTable} & \rotatebox{90}{Dog} & \rotatebox{90}{Horse} & \rotatebox{90}{Motorbike} & \rotatebox{90}{Person} & \rotatebox{90}{Pottedplant} & \rotatebox{90}{Sheep} & \rotatebox{90}{Sofa} & \rotatebox{90}{Train} & \rotatebox{90}{TV/Monitor}\\ \midrule
UniMatch \cite{yang2023revisiting}           & 75.2  & 93.6      &88.0     &67.0  &90.7  &72.0    &74.0  &93.2  &86.2  &93.9   &13.9  &90.9          &52.5  &90.1   &91.9       &80.7    &82.0          &54.7   &87.7  &22.3   &87.4 &66.3   \\ \midrule
UniMatch + Ours                  &\textbf{76.7}     &\textbf{94.0}       &\textbf{90.9}     &\textbf{67.8}  &\textbf{91.1}  &\textbf{73.7}    &\textbf{74.8}  &\textbf{93.3}  &\textbf{86.3}  &\textbf{94.2}   &\textbf{15.2}  &\textbf{94.2}          &\textbf{54.2}  &\textbf{91.1}   &\textbf{92.1}       &\textbf{81.5}    &\textbf{82.2}          &\textbf{60.9}   &\textbf{91.4}  &\textbf{25.7}   &\textbf{88.1} & \textbf{68.3}   \\
 \bottomrule
\end{tabular}}
\end{subtable}
\end{table*}
\section{Conclusion}
In this paper, we propose a novel approach that reduces the reliance on segmentation scores of the trained teacher model in pseudo-labeling unlabeled images. To do so, we propose a two-step approach. First, an ensemble of segmentation and detection models is used to identify reliable pseudo-labeled pixels. Second, a per-pixel weight is calculated to weigh the pseudo-labeled pixels. To determine this weight we first construct a prototype based on the labeled pixels and the reliable pseudo-labeled pixels identified from the first stage. Then the per-pixel weight is the similarity between the pixel and the prototype via rank--statistics. We show that our approach can be easily integrated into other approaches by integrating it into four  approaches, which improves their performance in all data partitions if two recognized segmentation datasets, Cityscapes and Pascal VOC.
\\
\textbf{Acknowledgement.} This work is supported by the National Science Foundation (IIS-2123920,  IIS-2212046).

\input{supplementary}
\newpage
\appendix
\bibliographystyle{splncs04}
\bibliography{main}
\end{document}

%% file: table/voc.tex
\begin{table*}[]
\caption{Quantitative results of different semi-supervised segmentation methods on Pascal VOC classic and blender set. We report Mean IoU  under various partition protocols and show the improvements ({\color[HTML]{6434FC} $\Delta$}) over corresponding baseline. 
}
\label{tab:voc}
\begin{subtable}{\linewidth}
\centering
\resizebox{0.99\linewidth}{!}{%
\begin{tabular}{lr|ccccc|ccc}
\hline
\multicolumn{2}{c|}{\multirow{2}{*}{Method}} & \multicolumn{5}{c|}{\textit{Classic}} & \multicolumn{3}{c}{\textit{Blender}} \\ \cline{3-10} & & 1/16  & 1/8   & 1/4   & 1/2   & Full  & 1/16 & 1/8 & 1/4 \\ \hline\hline
\multicolumn{10}{l}{\textbf{ResNet-50}} \\\hline
\textit{Supervised}  &   & 44.0  & 52.3  & 61.7  & 66.7  & 72.9  &- & - & - \\
PC$^2$Seg \cite{zhong2021pixel}  & {\color[HTML]{656565} {[}CVPR'21{]}}   & 56.9 & 64.6 & 67.6 & 70.9 & 72.3 & -  & - & - \\
PseudoSeg \cite{zou2020pseudoseg}  & {\color[HTML]{656565} {[}ICLR'21{]}}   & 56.9 & 64.6 & 67.6 & 70.9 & 72.2 & -  & - & - \\
ST++ \cite{yang2022st++}       & {\color[HTML]{656565} {[}CVPR'22{]}}  & -    & -    & -    & -    & -    & 72.6 & 74.4 & 75.4 \\\hline
AugSeg  \cite{zhao2023augmentation} & {\color[HTML]{656565} {[}CVPR'23{]}}  & 64.2 & 72.1 & 76.1 & 77.4 & 78.8 & 77.2 & 78.2 & 78.2 \\
AugSeg + Ours/{\color[HTML]{6434FC} $\Delta$} &  
& \textbf{66.4}/{\color[HTML]{6434FC} $2.2$} 
& \textbf{73.9}/{\color[HTML]{6434FC} $1.8$} 
& \textbf{77.6}/{\color[HTML]{6434FC} $1.5$} 
& \textbf{78.3}/{\color[HTML]{6434FC} $0.9$} 
& \textbf{79.3}/{\color[HTML]{6434FC} $0.5$} 
& \textbf{79.5}/{\color[HTML]{6434FC} $2.3$} 
& \textbf{79.1}/{\color[HTML]{6434FC} $0.9$} 
& \textbf{79.8}/{\color[HTML]{6434FC} $1.6$} \\\hline
AEL \cite{hu2021semi} & {\color[HTML]{656565} {[}NeurIPS'21{]}}  & 62.9 & 64.1 & 70.3 & 72.7 & 74.0 & 74.1 & 76.1 & 77.9 \\
AEL + Ours/{\color[HTML]{6434FC} $\Delta$} &  
& \textbf{66.1}/{\color[HTML]{6434FC} $3.2$}                
& \textbf{66.4}/{\color[HTML]{6434FC} $2.3$}                 
& \textbf{72.2} /{\color[HTML]{6434FC} $1.9$}                
& \textbf{74.3}/{\color[HTML]{6434FC} $1.6$}                 
& \textbf{74.9}/{\color[HTML]{6434FC} $0.9$}                 
& \textbf{77.0}/{\color[HTML]{6434FC} $2.9$}                 
& \textbf{78.1}/{\color[HTML]{6434FC} $2.0$}                 
& \textbf{79.2}/{\color[HTML]{6434FC} $1.3$}                                                 \\
\hline
% =======================
U2PL \cite{wang2022semi} & {\color[HTML]{656565} {[}CVPR'22{]}}  & 63.3 & 65.5 & 71.6 & 73.8 & 75.1 & 74.7 & 77.4 & 77.5 \\
U2PL  + Ours/{\color[HTML]{6434FC} $\Delta$} &  
& \textbf{66.0}/{\color[HTML]{6434FC} $2.7$}                 
& \textbf{67.6}/{\color[HTML]{6434FC} $2.1$}                 
& \textbf{73.2}/{\color[HTML]{6434FC} $1.6$}                 
& \textbf{75.5}/{\color[HTML]{6434FC} $1.7$}                 
& \textbf{75.9}/{\color[HTML]{6434FC} $0.8$}               
& \textbf{77.7}/{\color[HTML]{6434FC} $3.0$}                 
& \textbf{79.8}/{\color[HTML]{6434FC} $2.4$}               
& \textbf{79.4}/{\color[HTML]{6434FC} $1.9$}\\\hline
UniMatch \cite{yang2023revisiting} & {\color[HTML]{656565} {[}CVPR'23{]}}  & 71.9 & 72.5 & 76.0 & 77.4 & 78.7 & 78.1 & 79.0 & 79.1 \\
UniMatch + Ours/{\color[HTML]{6434FC} $\Delta$} &      
& \textbf{73.9}/{\color[HTML]{6434FC} $2.0$}                 
& \textbf{74.3}/{\color[HTML]{6434FC} $1.8$}                 
& \textbf{77.3}/{\color[HTML]{6434FC} $1.3$}                 
& \textbf{78.8}/{\color[HTML]{6434FC} $1.4$}                 
& \textbf{79.6}/{\color[HTML]{6434FC} $0.9$}                 
& \textbf{80.2}/{\color[HTML]{6434FC} $2.1$}                 
& \textbf{80.6}/{\color[HTML]{6434FC} $1.6$}                 
& \textbf{80.2}/{\color[HTML]{6434FC} $1.1$}\\
 \hline\hline
\multicolumn{10}{l}{\textbf{ResNet-101 }} \\\hline
\textit{Supervised} &   & 45.1 & 55.3 & 64.8 & 69.7 & 73.5 & 70.6 & 75.0 & 76.5 \\
CPS  \cite{chen2021semi}    & {\color[HTML]{656565} {[}CVPR'21{]}}  & 64.1 & 67.4 & 71.7 & 75.9 & - & 72.2 & 75.8 & 77.6 \\
PS-MT \cite{liu2022perturbed}       & {\color[HTML]{656565} {[}CVPR'22{]}}  & 65.8 & 69.6 & 76.6 & 78.4 & 80.0 & 75.5 & 78.2 & 78.7 \\
PCR \cite{xu2022semi}          & {\color[HTML]{656565} {[}NeurIPS'22{]}}  & 70.1 & 74.7 & 77.2 & 78.5 & 80.7 & 78.6 & 80.7 & 80.8 \\
DAW  \cite{sun2023daw}            & {\color[HTML]{656565} {[}NeurIPS'23{]}}  & 74.8 & 77.4 & 79.5 & 80.6 & 81.5 & 78.5 & 78.9 & 79.6 \\
CFCG  \cite{li2023cfcg}          & {\color[HTML]{656565} {[}ICCV'23{]}}  & - & - & - & - & - & 77.4 & 79.4 & 80.4 \\\hline
AugSeg \cite{zhao2023augmentation}     & {\color[HTML]{656565} {[}CVPR'23{]}}  & 71.1 & 75.5 & 78.8 & 80.3 & 81.4 & 79.3 & 81.5 & 80.5 \\
AugSeg + Ours/{\color[HTML]{6434FC} $\Delta$} &  
& \textbf{73.1}/{\color[HTML]{6434FC} $2.0$}                 
& \textbf{77.2}/{\color[HTML]{6434FC} $1.7$}                 
& \textbf{80.3}/{\color[HTML]{6434FC} $1.5$}                 
& \textbf{81.1}/{\color[HTML]{6434FC} $0.8$}                 
& \textbf{81.8}/{\color[HTML]{6434FC} $0.4$}                 
& \textbf{81.2}/{\color[HTML]{6434FC} $1.9$}                 
& \textbf{82.8}/{\color[HTML]{6434FC} $1.3$}                 
& \textbf{81.2}/{\color[HTML]{6434FC} $0.7$}\\ \hline
AEL \cite{hu2021semi} & {\color[HTML]{656565} {[}NeurIPS'21{]}}  & 66.1 & 68.3 & 71.9 & 74.4 & 78.9 & 77.2 & 77.6 & 78.1 \\
AEL+ Ours/{\color[HTML]{6434FC} $\Delta$} &    
&\textbf{69.8}/{\color[HTML]{6434FC} $3.7$} 
&\textbf{71.6}/{\color[HTML]{6434FC} $3.3$}   
&\textbf{74.0}/{\color[HTML]{6434FC} $2.1$}                  
& \textbf{76.1}/{\color[HTML]{6434FC} $1.7$}                 
& \textbf{80.3}/{\color[HTML]{6434FC} $1.4$}                 
& \textbf{80.5}/{\color[HTML]{6434FC} $3.3$}          
&\textbf{80.6}/{\color[HTML]{6434FC} $3.0$}                  
& \textbf{80.8}/{\color[HTML]{6434FC} $2.7$}\\ \hline
U2PL \cite{wang2022semi}      & {\color[HTML]{656565} {[}CVPR'22{]}}  & 68.0 & 69.2 & 73.7 & 76.2 & 79.5 & 77.2 & 79.0 & 79.3\\
U2PL  + Ours /{\color[HTML]{6434FC} $\Delta$} & 
&\textbf{71.1}/{\color[HTML]{6434FC} $3.1$}
& \textbf{72.0}/{\color[HTML]{6434FC} $2.8$}                                        
& \textbf{75.6}/{\color[HTML]{6434FC} $1.9$}                                        
&\textbf{78.0}/{\color[HTML]{6434FC} $1.8$}                                         
&\textbf{81.0}/{\color[HTML]{6434FC} $1.5$}                                         
&\textbf{80.1}/{\color[HTML]{6434FC} $2.9$}                                         
& \textbf{81.5}/{\color[HTML]{6434FC} $2.5$}                                        
&\textbf{81.6}/{\color[HTML]{6434FC} $2.3$}\\ \hline
UniMatch \cite{yang2023revisiting} & {\color[HTML]{656565} {[}CVPR'23{]}}  & 75.2 & 77.2 & 78.8 & 79.9 & 81.2 & 80.9 & 81.9 & 80.4 \\
UniMatch + Ours  /{\color[HTML]{6434FC} $\Delta$} & 
& \textbf{76.7} /{\color[HTML]{6434FC}$1.5$}                                       
& \textbf{78.5}/{\color[HTML]{6434FC} $1.3$}                                        
& \textbf{80.0}/{\color[HTML]{6434FC} $1.2$}                                        
& \textbf{80.9}/{\color[HTML]{6434FC} $1.0$}                                        
& \textbf{81.7}/{\color[HTML]{6434FC} $0.5$}                                        
& \textbf{83.0}/{\color[HTML]{6434FC} $2.1$}                                        
& \textbf{83.5}/{\color[HTML]{6434FC} $1.6$}                                        
& \textbf{81.4}/{\color[HTML]{6434FC} $1.0$}\\\hline
\end{tabular}}
\end{subtable}
\end{table*}

%% file: table/ct.tex
\begin{table*}[]
\centering
\caption{Quantitative results of different semi-supervised segmentation methods on the Cityscapes validation set. We report Mean IoU under various partition protocols and show the improvements ({\color[HTML]{6434FC} $\Delta$}) over the corresponding baseline.}
\label{tab:ct}
\resizebox{1\linewidth}{!}{%
\begin{tabular}{l r|cccc|cccc}
\hline
\multicolumn{2}{c|}{\multirow{2}{*}{Method}} & \multicolumn{4}{c|}{\textbf{ResNet-50}}                           & \multicolumn{4}{c}{\textbf{ResNet-101}}                          \\ \cline{3-10} 
\multicolumn{1}{c}{}            &            & 1/16    & 1/8     & 1/4     & 1/2    & 1/16    & 1/8     & 1/4     & 1/2    \\ \hline
\textit{Supervised}    &                & 63.34   & 68.73   & 74.14   & 76.62  & 66.3    & 72.8    & 75.0    & 78.0   \\
CPS \cite{chen2021semi}  & {\color[HTML]{656565} {[}CVPR'21{]}}                      & 69.79   & 74.39   & 76.85   & 78.64  & 69.8    & 74.3    & 74.6    & 76.8   \\
PS-MT \cite{liu2022perturbed}       & {\color[HTML]{656565} {[}CVPR'22{]}}           & -       & 75.76   & 76.92   & 77.64  & -       & 76.9    & 77.6    & 79.1   \\
PCR \cite{xu2022semi}    & {\color[HTML]{656565} {[}NeurIPS'22{]}}                      & -       & -       & -       & -      & 73.4    & 76.3    & 78.4    & 79.1   \\
CFCG \cite{li2023cfcg}   & {\color[HTML]{656565} {[}ICCV'23{]}}                      & 76.1       & 78.9       & 79.3       & 80.1      & 77.8    & 79.6    & 80.4    & 80.9   \\ \hline
AugSeg \cite{zhao2023augmentation}  & {\color[HTML]{656565} {[}CVPR'23{]}}          & 73.7    & 76.4    & 78.7    & 79.3   & 75.2    & 77.8    & 79.6    & 80.4   \\
AugSeg + Ours/{\color[HTML]{6434FC} $\Delta$}              &                  
& \textbf{76.0}/{\color[HTML]{6434FC} $2.3$} 
& \textbf{78.3}/{\color[HTML]{6434FC} $1.9$} 
& \textbf{80.2}/{\color[HTML]{6434FC} $1.5$} 
& \textbf{80.3}/{\color[HTML]{6434FC} $1.0$} 
& \textbf{77.3}/{\color[HTML]{6434FC} $2.1$} 
& \textbf{79.3}/{\color[HTML]{6434FC} $1.5$} 
& \textbf{81.4}/{\color[HTML]{6434FC} $1.8$} 
& \textbf{81.3}/{\color[HTML]{6434FC} $0.9$} \\ \hline
AEL \cite{hu2021semi}        & {\color[HTML]{656565} {[}NeurIPS'21{]}}                 & 68.2    & 72.7    & 74.9    & 77.5   & 74.5    & 75.6    & 77.5    & 79.0   \\
AEL + Ours/{\color[HTML]{6434FC} $\Delta$}                   &                
& \textbf{71.6}/{\color[HTML]{6434FC} $3.4$} 
& \textbf{75.4}/{\color[HTML]{6434FC} $2.7$} 
& \textbf{77.0}/{\color[HTML]{6434FC} $2.1$} 
& \textbf{79.9}/{\color[HTML]{6434FC} $2.4$} 
& \textbf{77.9}/{\color[HTML]{6434FC} $3.4$} 
& \textbf{78.8}/{\color[HTML]{6434FC} $3.2$} 
& \textbf{79.6}/{\color[HTML]{6434FC} $2.1$} 
& \textbf{80.1}/{\color[HTML]{6434FC} $1.1$} 
\\ \hline
U2PL \cite{wang2022semi}          & {\color[HTML]{656565} {[}CVPR'22{]}}            & 69.0    & 73.0    & 76.3    & 78.6   & 74.9    & 76.5    & 78.5    & 79.1   \\
U2PL + Ours/{\color[HTML]{6434FC} $\Delta$}             &                     
& \textbf{71.9}/{\color[HTML]{6434FC} $2.9$} 
& \textbf{75.8}/{\color[HTML]{6434FC} $2.8$} 
& \textbf{77.9}/{\color[HTML]{6434FC} $1.6$} 
& \textbf{79.9}/{\color[HTML]{6434FC} $1.3$} 
& \textbf{78.0}/{\color[HTML]{6434FC} $3.1$} 
& \textbf{79.5}/{\color[HTML]{6434FC} $3.0$} 
& \textbf{80.0}/{\color[HTML]{6434FC} $1.5$} 
& \textbf{79.7}/{\color[HTML]{6434FC} $0.6$} 
\\ \hline
UniMatch \cite{yang2023revisiting}   & {\color[HTML]{656565} {[}CVPR'23{]}}         & 75.0    & 76.8    & 77.5    & 78.6   & 76.6    & 77.9    & 79.2    & 79.5   \\
UniMatch + Ours/{\color[HTML]{6434FC} $\Delta$}           &                   
& \textbf{77.1}/{\color[HTML]{6434FC} $2.1$} 
& \textbf{78.5}/{\color[HTML]{6434FC} $1.7$} 
& \textbf{78.7}/{\color[HTML]{6434FC} $1.2$} 
& \textbf{79.3}/{\color[HTML]{6434FC} $0.7$} 
& \textbf{78.4}/{\color[HTML]{6434FC} $1.8$} 
& \textbf{79.6}/{\color[HTML]{6434FC} $1.7$} 
& \textbf{80.5}/{\color[HTML]{6434FC} $1.3$} 
& \textbf{80.7}/{\color[HTML]{6434FC} $1.2$} 
\\ \hline
\end{tabular}%
}
\end{table*}

%% file: supplementary.tex
\setlist[itemize]{noitemsep,leftmargin=*,topsep=0em}
\setlist[enumerate]{noitemsep,leftmargin=*,topsep=0em}
%\begin{document}

\title{Weighting Pseudo-Labels via High-Activation Feature Index Similarity and Object Detection for Semi-Supervised Segmentation\\
\textit{Supplementary Material}} 
\titlerunning{Weighting Pseudo-Labels}


\author{ }

\authorrunning{P. Howlader et al.}

% \institute{Stonybrook University \and
% \email{\{phowlader; samaras\}@cs.stonybrook.edu}\\

\institute{ 
}
\maketitle
\textbf{Summary:}
We provide additional analyses and results of our method, including:
\begin{itemize}
    % \item \textbf{Per-pixel Learning Weight Visualization}
    \item Obtaining Pseudo Object Detection Training Data from Segmentation Masks
    \item The Effect of the Object Detector in Improving Pseudo Label Accuracy
    \item Analysis of pseudo-labeling accuracy
    \item Analysis of training hyper-parameters
    \item Memory bank and the effect of its memory size
    \item Analysis of top-rank indices of class prototypes throughout training
    \item Comparison of pseudo-labeling accuracy with Euclidean distance
    \item Visualizing the top-ranked features
    \item Adapting our method to Transformer-based models
    \item Evaluation of performance on MS COCO
   % \item Effect of Memory Size
    \item Qualitative results
\end{itemize}

\section{Obtaining Pseudo Object Detection Training Data from Segmentation Masks}\label{additionalcityscapestrain}
We train the object detector from scratch using only the labeled segmentation data. 
Given an image and its semantic segmentation mask, we first separate the mask of each category into separate connected components. For each component, we extract the smallest bounding box containing it to use as pseudo-training data for the object detector. Apparently, each box might contain more than one object instance. Thus, our object detector is essentially trained to detect bounding boxes containing groups of pixels belonging to the same category rather than bounding boxes containing single object. This suffices for semantic segmentation tasks such as ours where instance differentiation is not necessary. 

Further, we discard bounding boxes of background classes such as \textit{sky}, \textit{vegetation}, and \textit{buildings} since their bounding boxes tend to cover almost the entire image. Specifically, for the Pascal VOC dataset, we generate bounding boxes for all classes except for ``\textit{background}''. For the Cityscapes dataset, we generate bounding boxes for the foreground classes: Person, Rider, Car, Bicycle, Motorcycle, Train, Truck, Bus, Traffic Light, and Traffic Sign.\\  

\section{The Effect of the Object Detector in Improving Pseudo Label Accuracy}
The detection model is used in an ensemble with a segmentation model to identify reliable pseudo-label pixels. Here we analyze the performance of the detection model in limited labeled data scenarios and also how it improves the reliability of pseudo-labeled pixels when using together with the segmentation network. For our analysis, we use a Deeplabv3+ segmentation model with ResNet-101 backbone\cite{chen2018encoder} and a Faster R-CNN object model. Both are trained on the $\frac{1}{16}$ split of the Pascal VOC dataset (\textit{Classic}) containing $92$ labeled images. 

The performance of the Faster R-CNN is shown in the first row in Table \ref{tab:bboxperformance}, denoted as \textbf{{Faster R-CNN}}. We report the box-level accuracy for all detected bounding boxes with confidence scores $\ge 0.85$. A detected box is considered ``correct'' if its IoU $\ge 0.8$ with a ground truth bounding box (discussed in Sec. \ref{additionalcityscapestrain}) and ``incorrect'' otherwise. As can be seen, bounding boxes predicted by this Faster R-CNN model are fairly accurate: above 80\% in all cases, even with only $92$ training images.

More importantly, we show that this object detector improves the pseudo-label accuracy when used together with the segmentation network. We consider three group of pixels:  1) a baseline pseudo-labeling method that select pixels with high confidence scores from the segmentation model (denoted as ``\textbf{Pixel}'' in Table \ref{tab:bboxperformance}), 2) all pixels in the first group that are labeled as the same class by the object detector ( denoted as ``\textbf{Pixel $\cap$ BBOX}'' in Table \ref{tab:bboxperformance}) and 3) all pixels in the first group that do not intersect with any detected bounding boxes of the same category (denoted as \textbf{``Pixel $\setminus$ BBOX''} in Table \ref{tab:bboxperformance}).
As can be seen, pixels that are labeled as the same classes by both models are more reliable than just using the segmentation model.

\begin{table*}[]
\caption{Analysis of the performance of Faster R-CNN trained on limited labeled data and pseudo-label accuracy of pixels based on an ensemble of object detector and segmentation model on $\frac{1}{16}$ Pascal (\textit{Classic}) set. The first row reports the bounding box accuracy of Faster R-CNN. Pixel denotes the pseudo-labels by Deeplabv3+(ResNet-101) \cite{chen2018encoder}, BBox denotes bounding boxes generated by Faster R-CNN \cite{ren2015faster}.  }
\label{tab:bboxperformance}
\begin{subtable}{1.0\linewidth}
\centering
\resizebox{1.0\linewidth}{!}{
\begin{tabular}{@{}c|ccccccccccccccccccccc@{}}
\toprule
Category                  & \rotatebox{90}{Background} & \rotatebox{90}{Aeroplane} & \rotatebox{90}{Bicycle} & \rotatebox{90}{Bird} & \rotatebox{90}{Boat} & \rotatebox{90}{Bottle}  & \rotatebox{90}{Bus} & \rotatebox{90}{Car} & \rotatebox{90}{Cat} & \rotatebox{90}{Chair} & \rotatebox{90}{Cow} & \rotatebox{90}{DiningTable} & \rotatebox{90}{Dog} & \rotatebox{90}{Horse} & \rotatebox{90}{Motorbike} & \rotatebox{90}{Person} & \rotatebox{90}{Pottedplant} & \rotatebox{90}{Sheep} & \rotatebox{90}{Sofa} & \rotatebox{90}{Train} & \rotatebox{90}{TV/Monitor}\\ \midrule
Faster R-CNN            & -  & 83.7      & 89.6    & 80.4 & 83.9 & 87.5   & 82.3 & 90.1 & 92.6 & 80.1  & 92.3 & 80.4         & 94.7 & 93.9  & 80.7      & 93.6   & 81.3         & 82.5  & 80.8 & 83.1  & 84.7  \\ \midrule
Pixel                  & 80.2    & 71.9      & 55.1    & 83.4 & 79.7 & 73.3   & 68.2 & 87.4 & 82.2 & 32.8  & 80.2 & 46.5         & 70.9 & 85.4  & 74.7      & 57.4   & 45.1         & 72.0  & 40.3 & 67.1  & 79.6  \\
Pixel$\cap$BBOX          & -        & \textbf{78.6}      & \textbf{60.2}    & \textbf{88.5} & \textbf{83.9} & \textbf{79.2}   & \textbf{71.1} & \textbf{88.1} & \textbf{87.4} & \textbf{41.7}  & \textbf{83.5} & \textbf{52.9}         & \textbf{77.6} & \textbf{87.1}  & \textbf{76.6}      & \textbf{62.6}   & \textbf{57.9}         & \textbf{79.3}  & \textbf{59.6} & \textbf{75.6}  & \textbf{84.8}  \\
Pixel$\setminus$BBOX           & -  & 54.1      & 43.3    & 61.2 & 60.1 & 71.4   & 37.2 & 75.6 & 60.1 & 6.8   & 73.1 & 13.3         & 53.0 & 74.5  & 56.0      & 36.2   & 12.7         & 56.8  & 11.1 & 21.4  & 70.5  \\ \bottomrule
\end{tabular}}
\end{subtable}

\end{table*}

\section{Analysis of Pseudo-label Accuracy}\label{ablation_pseudolabelacc}
\begin{figure}[ht!]
    \centering
   \includegraphics[width=0.9\linewidth]{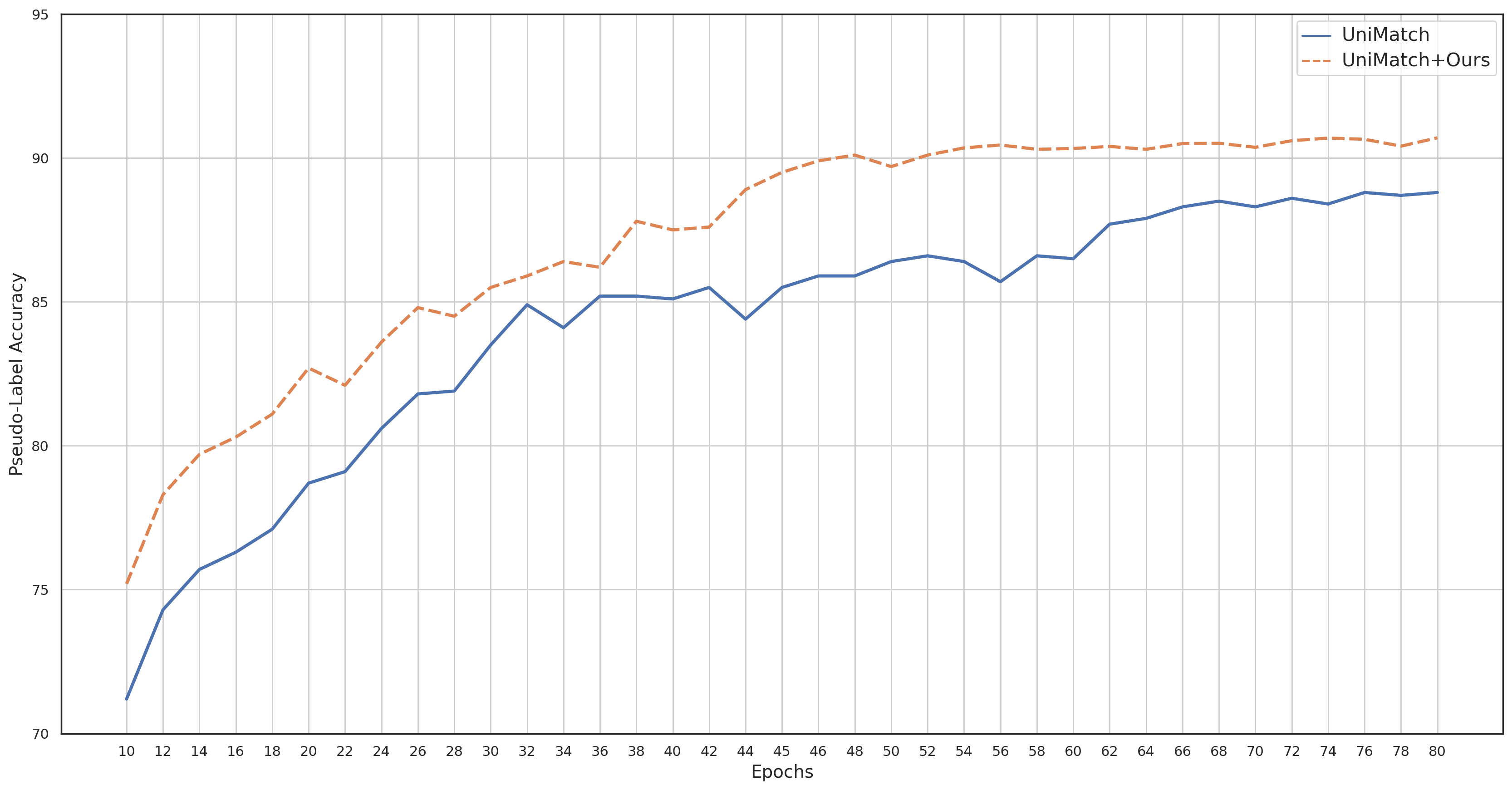} 
\caption{Pseudo-labeling accuracy in Pascal VOC unlabeled images
 }
  \label{fig:pseudolabelingaccunimatch}
\end{figure}
We analyze how our method, when integrated into UniMatch \cite{yang2023revisiting}, can improve the pseudo-label accuracy of this method. We train a vanilla UniMatch model and a UniMatch model with out method with 1/16 data partition of the Pascal VOC dataset (\textit{Classic}). As shown in \cref{fig:pseudolabelingaccunimatch}, our method pseudo-label pixels with up to $90\%$ accuracy and is consistently more accurate than UniMatch throughout the whole training process.

\section{Analysis of Hyper-Parameters}
In this section, we analyze the effects of different training hyper-parameters used by our method. We integrate our method into UniMatch \cite{yang2023revisiting} and train with 1/16 and 1/8 partition protocols of the Pascal VOC dataset (\textit{Classic})
\subsection{Analysis of the hyper-parameter $\alpha$ of semi-supervised segmentation loss}\label{ablation_alpha}
We analyze how our approach performs with different values of $\alpha$, which is used to balance the supervised loss and the unsupervised loss. The results of our approach are in Table \ref{tab:tunable_paramter_alpha}. %We observed $\alpha = 0.4$ as the optimal value achieving the best performance in the validation set. Thus, we apply this constant value across all our experiments.\\
\begin{table}[hbt!]
  \caption{Analysis of ($\alpha$) of semi-supervised segmentation loss (1/16 and 1/8 partition protocols of Pascal VOC Dataset) }
\label{tab:tunable_paramter_alpha}
%\small
  \centering
  \begin{tabular}{c|cc}
    \toprule
    %\multicolumn{2}{c}{Part}                   \\
    %\cmidrule(r){1-2}
   
$\alpha$ & 1/16 & 1/8 \\ 
 \midrule
   % \hline
   0 & 45.1 & 55.3 \\
0.2 & 75.1 & 78.0 \\
0.4 & \textbf{76.7} & \textbf{78.5}\\
0.6 & 76.3 & 78.1 \\
0.8 & 75.5 & 77.8\\

    \bottomrule
  \end{tabular}

\end{table}

\subsection{Analysis of Bounding Box Confidence Thresholds}\label{ablation_pseudolabelacc}
We analyze how our method performs with different bounding box confidence thresholds. As shown in Table \ref{tab:bboxconf}, the performance decreases with a very high confidence threshold. This is because the number of bounding boxes drastically reduces when increasing the threshold.  
\begin{table}[hbt!]
 \centering
 \caption{Analysis of BBox confidence threshold (1/16 and 1/8 partition protocols of Pascal VOC Dataset) }\label{tab:bboxconf}

\begin{tabular}{c|cc}
\toprule
BBox Confidence & 1/16          & 1/8           \\ \hline
0.80            & 76.1          & 78.3          \\
0.85            & \textbf{76.7} & \textbf{78.5} \\
0.90            & 75.9          & 77.4          \\
0.95            & 75.7          & 77.0          \\ \bottomrule
\end{tabular}

\end{table}

\section{Memory Bank Implementation Details} 
We use a memory bank to store features of labeled pixels and reliable pseudo-label pixels in each iteration, which is used to construct the per-class prototype. Since available memory is limited, only a random subset of features per class are included in the memory bank. Performing random sampling of the features to update the memory during training leads to a more diverse set of features per class. The memory follows First In First Out (FIFO) queue per class for computation and time efficiency \cite{alonso2021semi}. This helps in maintaining recent high-quality feature vectors.

\subsection{Effect of Memory Size}
The effect of the memory bank size (per-class) is studied in Table \ref{tab:memsize}. We observe that higher memory size leads to better performance, although from $256$ the performance tends to stabilise.
Since all elements from the memory bank are
used during the prototype generation, the computation
and memory complexity increases with a larger memory
bank, we selected a size of $256$ as a good tradeoff.

\begin{table}[]
\centering
\caption{Effect of our memory bank size (features per-class) $\psi$.  (1/16 partition protocol of Pascal VOC Dataset)}
\label{tab:memsize}
\begin{tabular}{@{}c|ccccc@{}}
\toprule
 $\psi$    & 32   & 64   & 128  & 256  & 512  \\ \midrule
mIoU & 74.6 & 75.1 & 75.9 & \textbf{76.7} & 76.2 \\ \bottomrule
\end{tabular}

\end{table}

\begin{figure}
    \centering
    \begin{minipage}[b]{.47\linewidth}
        \centering
        \includegraphics[width=0.98\linewidth]{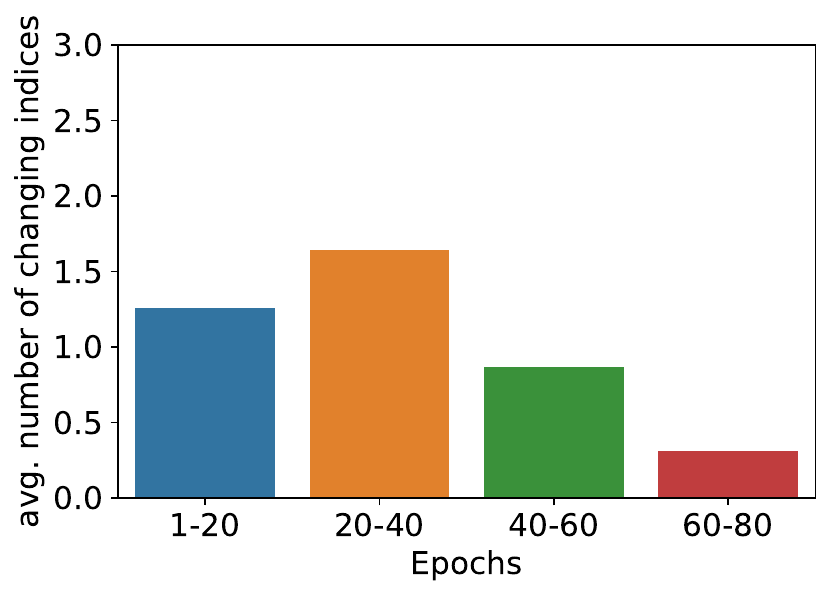}
        \captionof{figure}{Analysis of top-rank indices of class prototypes}
        \label{fig:prototypevarpic}
    \end{minipage}\hfill
    \begin{minipage}[b]{.47\linewidth}
    \centering
        \includegraphics[width=1.0\linewidth]{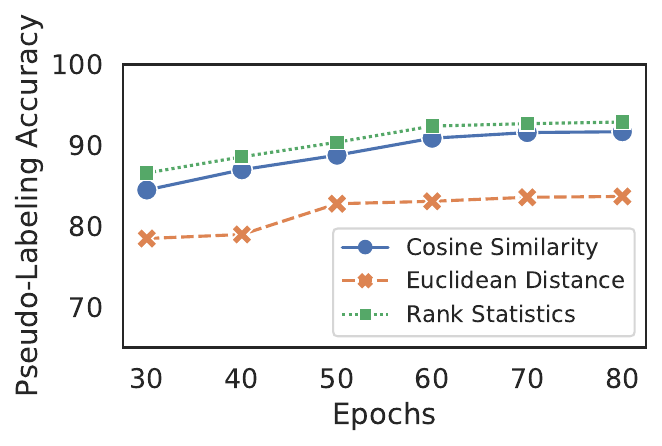}
        \captionof{figure}{Pseudo-labeling accuracy with euclidean distance}
        \label{fig:euclideanaccuracy}
    \end{minipage}
\end{figure}

\section{Visualizing the top-ranked features}
\begin{figure*}
\centering
\includegraphics[width=0.95\linewidth,]{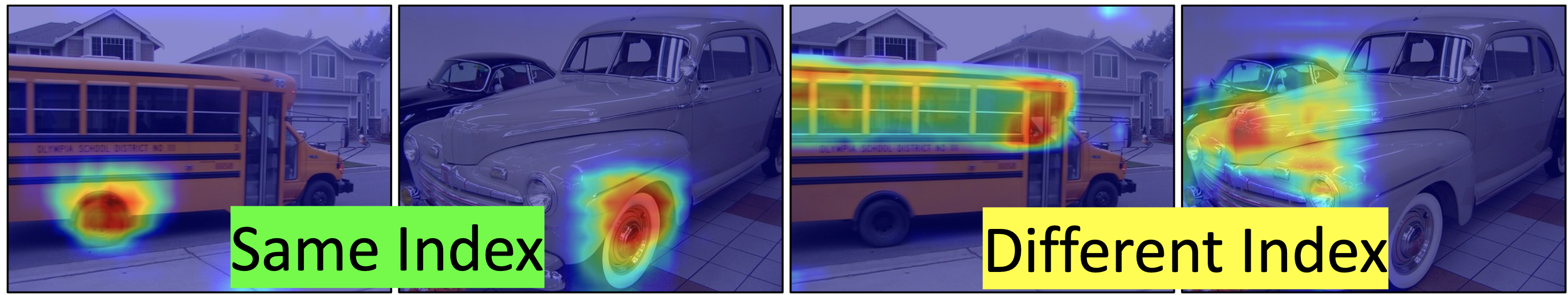}
% \vspace{0.5cm}
\caption{Visualizing the top-ranked (TR) features   
}
  \label{fig:TRfeaturepic}
\end{figure*}

Top-ranked (TR) features  often highlight discriminative object parts. In \cref{fig:TRfeaturepic} we show gradcam visualization of a common TR feature  between ``car" and ``bus" in the first two images and then show two different TR features, exclusive for each class, to illustrate the distinct regions each feature focuses on.

\section{ Adapting our method to Transformer-based models}
We analyse how our method improves the transformer-based models.  We experiment with SemiVL \cite{hoyer2023semivl} on Cityscapes and show the results in the Table \ref{tab:semivl}. We use the image embeddings prior to similarity map generation, to generate class prototypes and incorporate our per-pixel learning weight to the CLIP guidance loss.  
\begin{table}[]
\centering
\caption{Quantitative results of  on
the Cityscapes  dataset}
\label{tab:semivl}
\begin{tabular}{lc|cccc}
\multicolumn{1}{l|}{Method}  & Net  & 1/30   & 1/16  & 1/8     & 1/2   \\ \hline
\multicolumn{1}{l|}{SemiVL \cite{hoyer2023semivl}}  & Vit-B/16              & 76.2                     & 77.9          & 79.4                     & 80.6          \\
\multicolumn{1}{l|}{SemiVL \cite{hoyer2023semivl} + Ours} & Vit-B/16   & \textbf{77.5}            & \textbf{79.1}  &  \textbf{80.4}            &   \textbf{81.4} \\ \hline
\end{tabular}
\end{table}

\section{Evaluation of performance on  MS COCO}
In Table \ref{tab:mscoco}, when using our method on UniMatch \cite{yang2023revisiting} (Xception-65) on MS COCO. We observe that performance improves across all data splits. 
\begin{table}[]
\centering
\caption{Quantitative results of  on
the MS COCO  dataset}
\label{tab:mscoco}
\begin{tabular}{lccccc}
\multicolumn{1}{l|}{Method}  & 1/512  & 1/256   & 1/128  & 1/64     & 1/32   \\ \hline
\multicolumn{1}{l|}{UniMatch \cite{yang2023revisiting}}  & 31.9              & 38.9                     & 44.4          & 48.2                     & 49.8          \\
\multicolumn{1}{l|}{UniMatch \cite{yang2023revisiting} + Ours} & \textbf{33.7}   & \textbf{40.5}            & \textbf{46.6}  &  \textbf{50.3}            &   \textbf{51.7} \\ \hline
\end{tabular}
\end{table}

\section{Qualitative results}
\begin{figure*}
\centering
\includegraphics[width=0.95\linewidth,]{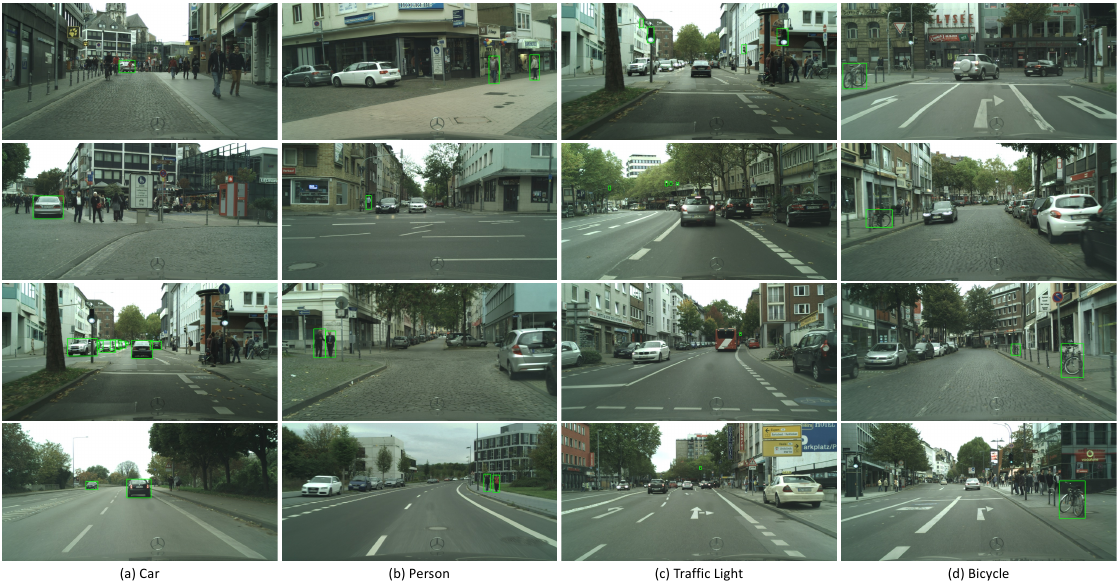}
% \vspace{0.5cm}
\caption{\textbf{Detection results on Cityscapes unlabeled images}  Faster R-CNN model trained on 1/16 labeled data in Cityscapes dataset and confidence $\ge 0.9$. From left to right: Car, Person, Traffic Light, Bicycle. 
}
  \label{fig:detections_1}
\end{figure*}

\begin{figure*}
\centering
\includegraphics[width=1\linewidth,]{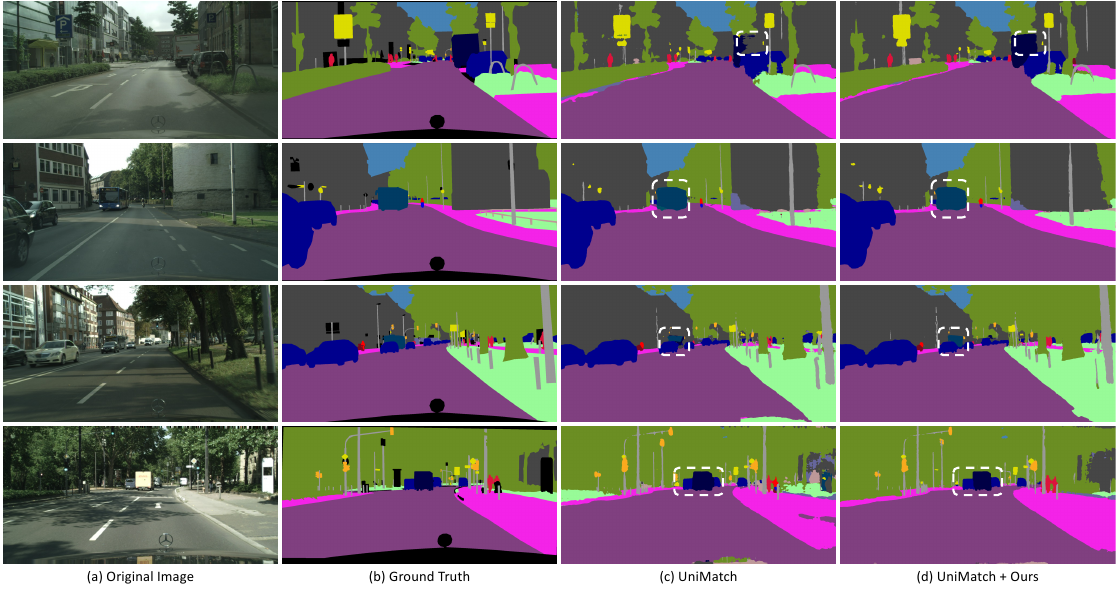}
% \vspace{0.5cm}
\caption{\textbf{Qualitative Results on Cityscapes dataset:}  (a) original image, (b) ground truth, (c) segmentations generated by UniMatch \cite{yang2023revisiting} compared to (d) which are segmentations generated when our method is integrated to UniMatch. The white boxes show the areas where our method improves the baseline \cite{yang2023revisiting}.  
}
  \label{fig:unimatchsupcitypic}
\end{figure*}

\begin{figure*}
\centering
\includegraphics[width=1\linewidth,]{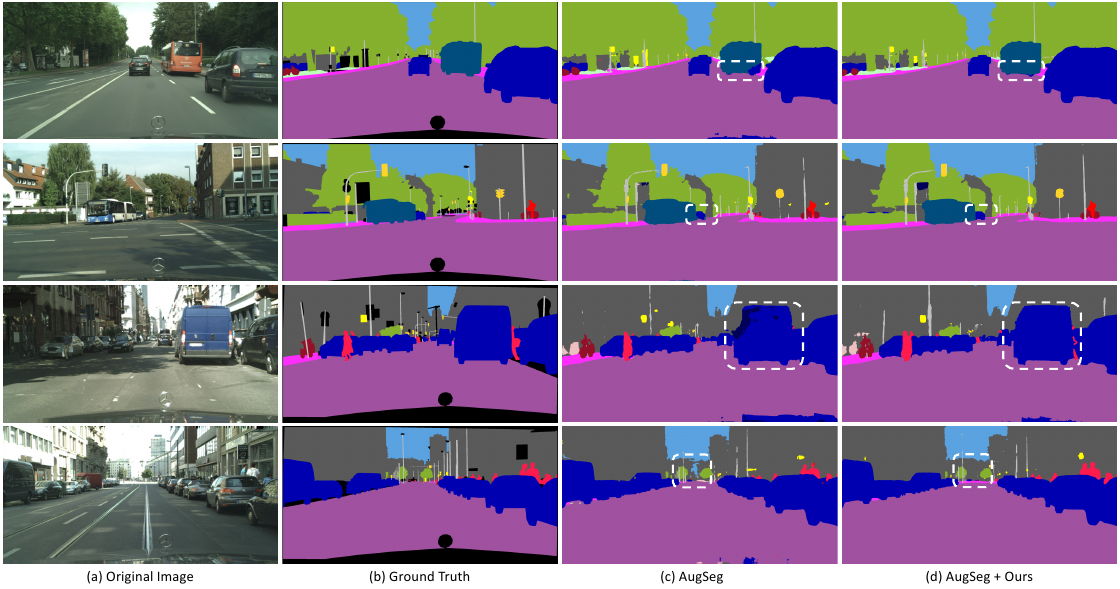}
% \vspace{0.5cm}
\caption{\textbf{Qualitative Results on Cityscapes dataset:} (a) original image, (b) ground truth, (c) segmentations generated by AugSeg \cite{zhao2023augmentation} compared to (d) which are segmentations generated when our method is integrated to AugSeg. The white boxes show the areas where our method improves the baseline \cite{zhao2023augmentation}.  
}
  \label{fig:augsegsupcitypic}
\end{figure*}

\begin{figure*}
\centering
\includegraphics[width=0.95\linewidth,]{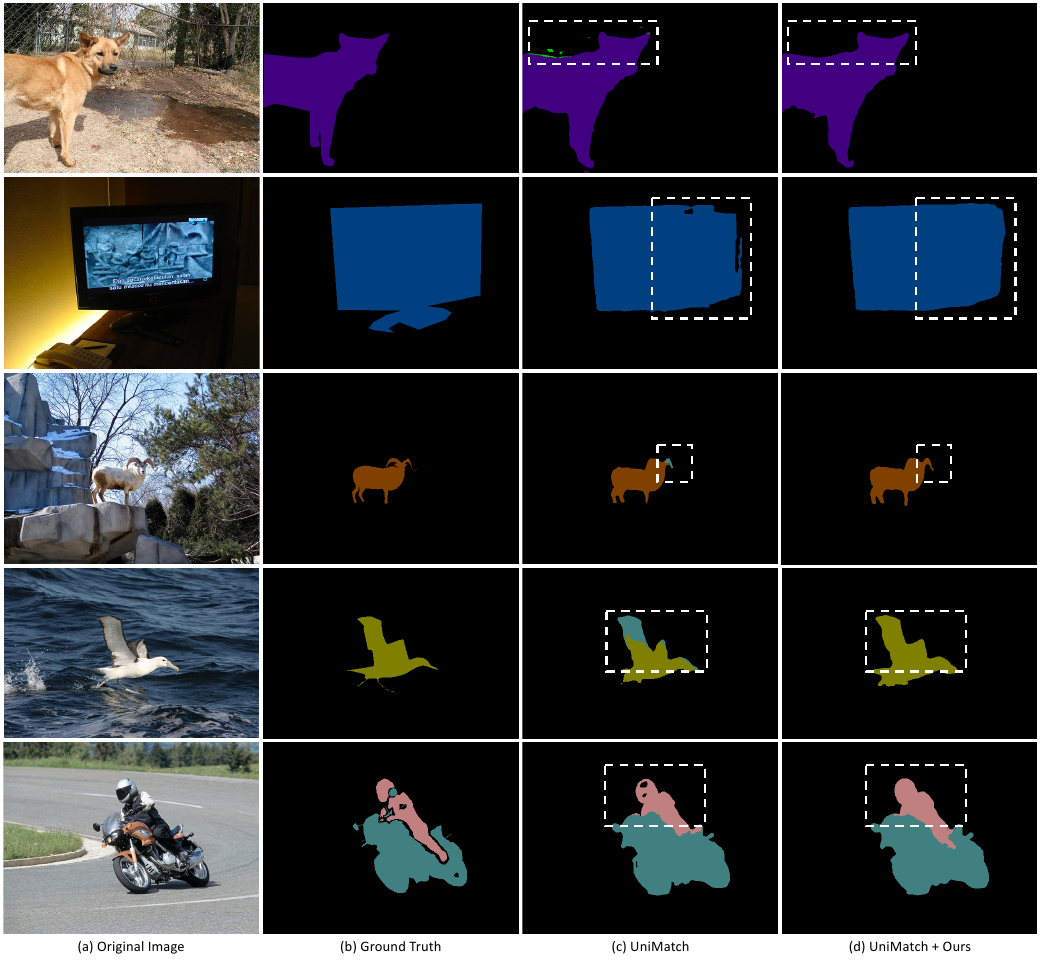}
% \vspace{0.5cm}
\caption{\textbf{Qualitative Results on Pascal dataset:} (a) original image, (b) ground truth, (c) segmentations generated by UniMatch \cite{yang2023revisiting} compared to (d) which are segmentations generated when our method is integrated to UniMatch. The white boxes show the areas where our method improves the baseline \cite{yang2023revisiting}.  
}
  \label{fig:unimatchsuppascalpic}
\end{figure*}

\begin{figure*}
\centering
\includegraphics[width=0.95\linewidth,]{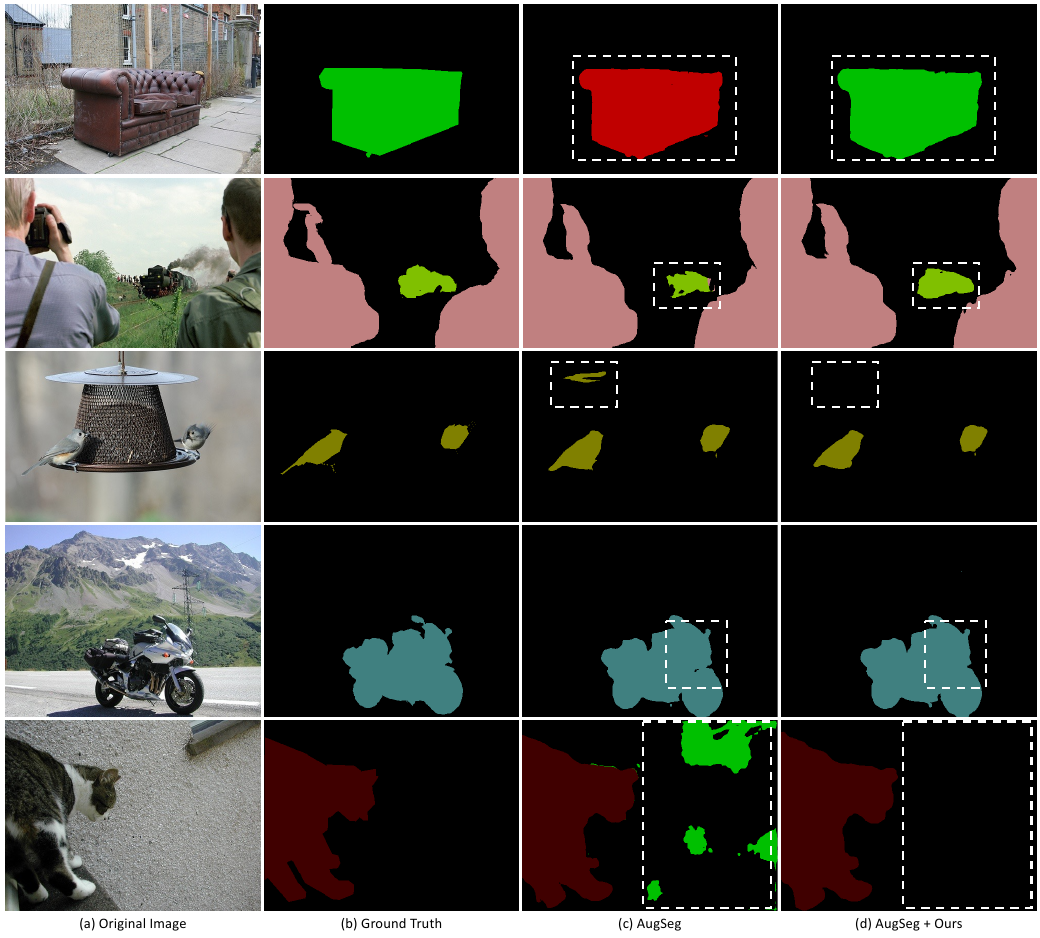}
% \vspace{0.5cm}
\caption{\textbf{Qualitative Results on Pascal dataset:} (a) original image, (b) ground truth, (c) segmentations generated by AugSeg \cite{zhao2023augmentation} compared to (d) which are segmentations generated when our method is integrated to AugSeg. The white boxes show the areas where our method improves the baseline \cite{zhao2023augmentation}.  
}
  \label{fig:augsegsuppascalpic}
\end{figure*}

\begin{figure*}
\centering
\includegraphics[width=0.95\linewidth,]{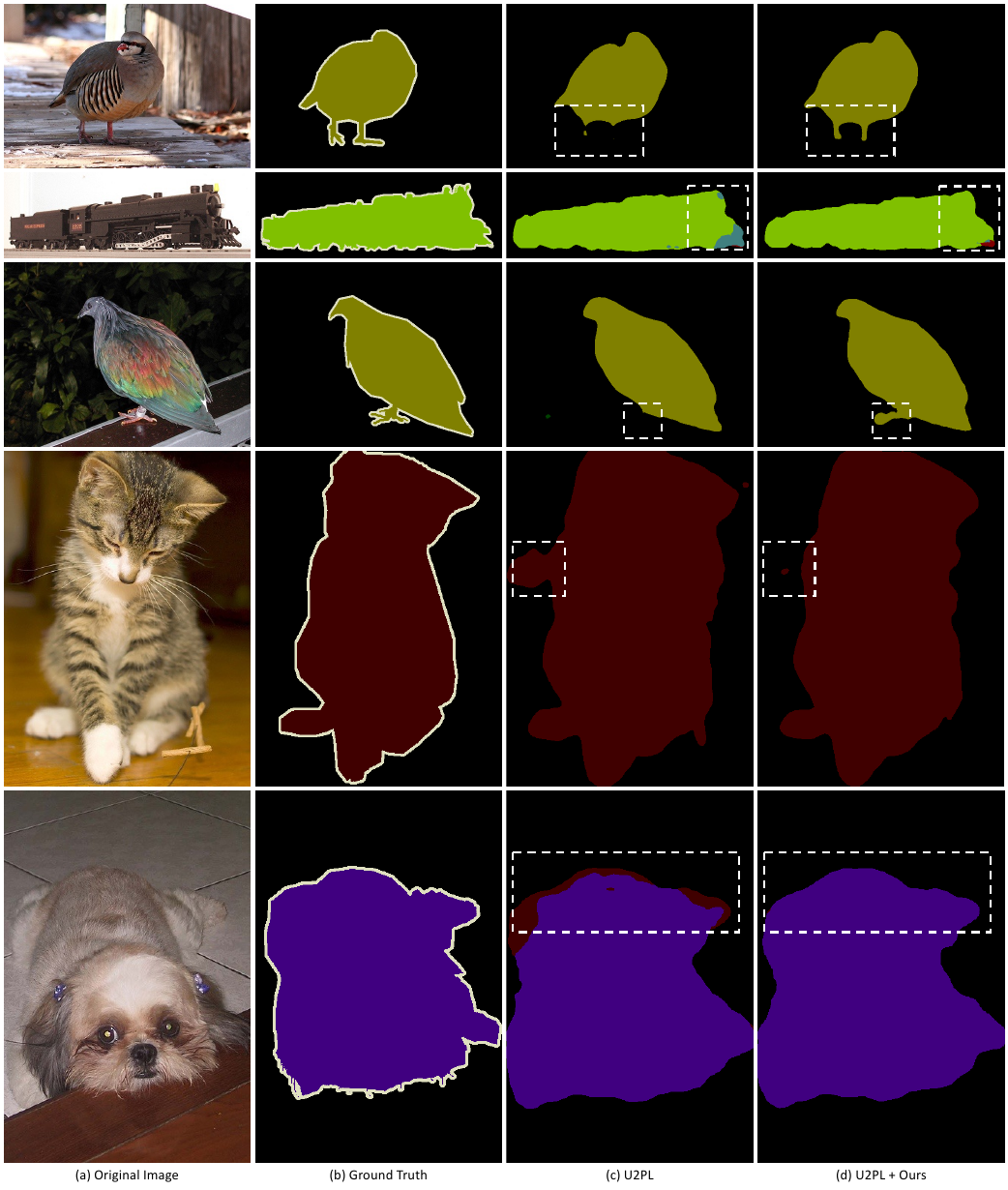}
% \vspace{0.5cm}
\caption{\textbf{Qualitative Results on Pascal dataset:}  (a) original image, (b) ground truth, (c) segmentations generated by U2PL \cite{wang2022semi} compared to (d) which are segmentations generated when our method is integrated to U2PL. The white boxes show the areas where our method improves the baseline \cite{wang2022semi}.  
}
  \label{fig:u2plsuppascalpic}
\end{figure*}

\begin{figure*}
\centering
\includegraphics[width=0.95\linewidth,]{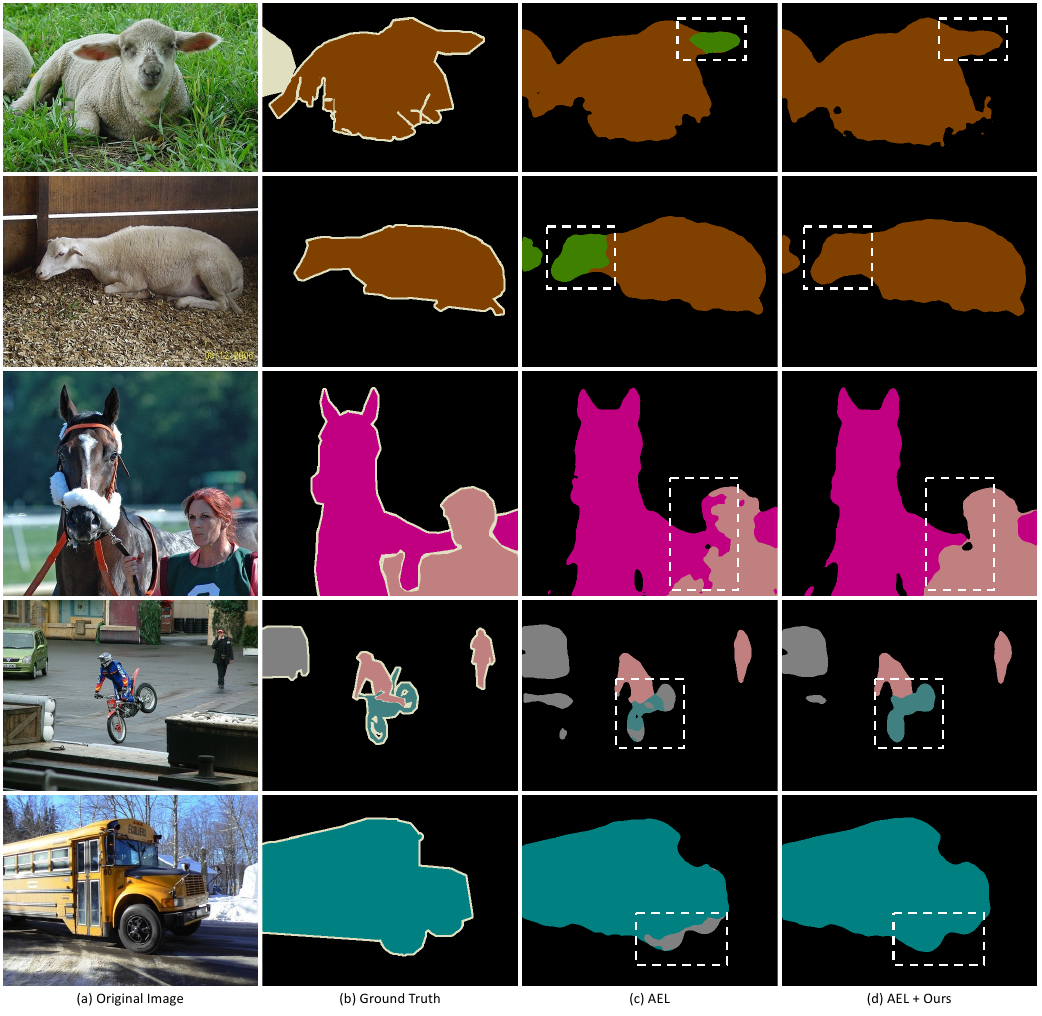}
% \vspace{0.5cm}
\caption{\textbf{Qualitative Results on Pascal dataset:} (a) original image, (b) ground truth, (c) segmentations generated by AEL \cite{hu2021semi} compared to (d) which are segmentations generated when our method is integrated to AEL. The white boxes show the areas where our method improves the baseline \cite{hu2021semi}.  
}
  \label{fig:aelsuppascalpic}
\end{figure*}

\begin{itemize}
\item \textbf{Detections}: In \cref{fig:detections_1}, we train a Faster R-CNN on $\frac{1}{16}$ of Cityscapes labeled data and visualize the detection results (confidence $\ge 0.9$) on Cityscapes unlabeled images. It can be observed that the detection boxes are relatively accurate when trained on limited labeled data.
\item \textbf{Pascal VOC}: In \cref{fig:unimatchsuppascalpic}, \cref{fig:augsegsuppascalpic}, \cref{fig:u2plsuppascalpic} and \cref{fig:aelsuppascalpic} we compare our method integrated into UniMatch \cite{yang2023revisiting}, AugSeg \cite{zhao2023augmentation}, U2PL \cite{wang2022semi} and AEL \cite{hu2021semi} respectively, with the corresponding baselines (UniMatch, AugSeg, U2PL and AEL).
 The visualization of the segmentation results indicate that our method improves the segmentation performance of all four baselines: UniMatch, AugSeg, U2PL and AEL. 

Note, all methods have been trained on $\frac{1}{16}$ data partition of Pascal VOC dataset (\textit{Classic}) and all visualizations are on Pascal VOC validation set. 
\item \textbf{Cityscapes}: In \cref{fig:unimatchsupcitypic}, \cref{fig:augsegsupcitypic}, we compare our method integrated into UniMatch \cite{yang2023revisiting} and AugSeg \cite{zhao2023augmentation} respectively, with the corresponding baselines (UniMatch and AugSeg).  The visualization of the segmentation results indicate that our method improves the segmentation performance of both baselines: UniMatch and AugSeg. 

Note, all methods have been trained on $\frac{1}{16}$ data partition of the Cityscapes dataset, and all visualizations are on the Cityscapes validation set.
\end{itemize}

%%%%%%%%%%%%%%%%%%%%%%%%
%%%%%%%%%%%%%%%%%%%%%%%%%%%%%%%%%%%%%%%%%%%%%%%%


\newpage
% \bibliography{main}
% \end{document}